\documentclass{article}

\PassOptionsToPackage{numbers, compress}{natbib}

\usepackage[preprint]{neurips_2026}

\usepackage[utf8]{inputenc} 
\usepackage[T1]{fontenc}    
\usepackage{url}            
\usepackage{booktabs}       
\usepackage{amsfonts}       
\usepackage{nicefrac}       
\usepackage{microtype}      

\usepackage[table]{xcolor}

\usepackage{amsmath}
\usepackage{amssymb}
\usepackage{mathtools}
\usepackage{amsthm}

\usepackage{graphicx}
\usepackage{makecell}
\usepackage{multirow}
\usepackage{subcaption}
\usepackage{wrapfig}
\usepackage{tabularx}
\usepackage{caption}
\usepackage{xurl} 
\usepackage{enumitem}
\usepackage{amssymb}
\usepackage{pifont}
\newcommand{\cmark}{\ding{51}}%
\newcommand{\xmark}{\ding{55}}%

\usepackage{hyperref}
\usepackage[capitalize,noabbrev]{cleveref} 

\usepackage{algorithm}
\usepackage{algorithmic}

\definecolor{darkblue}{rgb}{0.0, 0.0, 0.8}
\definecolor{darkgreen}{rgb}{0.0, 0.8, 0.0}
\definecolor{darkyellow}{rgb}{0.85, 0.65, 0.13}
\definecolor{darkred}{rgb}{0.8, 0.0, 0.0}
\ifodd 0    

\newcommand{\XR}[1]{{\color{darkgreen} #1}}

\newcommand{\rev}[1]{{\color{blue}#1}}

\else
\newcommand{\rev}[1]{#1}

\newcommand{\XR}[1]{#1}

\fi 

\ifodd 0
\newcommand{\rebuttal}[1]{{\color{blue}#1}}
\else
\newcommand{\rebuttal}[1]{#1}
\fi

\ifodd 1    

\newcommand{\XRii}[1]{{\color{black} #1}} 

\else
\newcommand{\XRii}[1]{#1}

\fi 

\ifodd 0    

\newcommand{\XRiii}[1]{{\color{darkgreen} #1}}

\else
\newcommand{\XRiii}[1]{#1}

\fi 
\title{More Than Meets the Eye? Uncovering the Reasoning-Planning Disconnect in Training Vision-Language Driving Models}

\author{%
  \textbf{Xurui Song}$^{1,2}$\thanks{Equal contribution.} \quad
  \textbf{Shuo Huai}$^{1,}$\footnotemark[1] \quad
  \textbf{Jingjing Jiang}$^{1}$ \quad
  \textbf{Jiayi Kong}$^{1,2}$ \quad
  \textbf{Jun Luo}$^{1,}$\thanks{Corresponding author.} \\[1.2ex]
  $^1$College of Computing and Data Science, Nanyang Technological University, Singapore \\
  $^2$S-Lab, Nanyang Technological University, Singapore
}

\begin{document}

\maketitle

\begingroup
\renewcommand\thefootnote{}%
\footnotetext{Code and dataset: \url{https://github.com/SONG0257/DriveMind}}%
\endgroup

\begin{abstract}
Vision-Language Model (VLM) driving agents promise explainable end-to-end autonomy by first producing natural-language reasoning and then predicting trajectory planning.
However, whether planning is \textbf{causally} driven by this reasoning remains a critical but unverified assumption. 
To investigate this, we build DriveMind, a large-scale driving Visual Question Answering corpus with plan-aligned Chain-of-Thought (CoT), automatically generated from nuPlan. 
Our 
data generation process
converts sensors and annotations into structured inputs and, crucially, separates priors from to-be-reasoned signals, enabling clean information ablations.
Using DriveMind, we train representative VLM agents with Supervised Fine-Tuning and Group Relative Policy Optimization and evaluate them with nuPlan’s metrics.
Our results, unfortunately, indicate a consistent \textbf{causal disconnect} in reasoning-planning: removing ego/navigation priors causes large drops in planning scores, whereas removing CoT produces only minor changes. {Attention analysis further shows that planning primarily focuses on priors rather than the CoT}.
\rev{Based on this evidence, we propose the Reasoning-Planning Decoupling Hypothesis, 
positing that the training-yielded reasoning is an ancillary byproduct rather than a causal mediator.}
To enable efficient
diagnosis, we 
introduce a novel, training-free probe that measures an agent's reliance on priors by evaluating its planning robustness against minor input perturbations. 
%
In summary, we provide the community with a new dataset
and
a diagnostic tool to evaluate the \textbf{causal fidelity} of future models.
\end{abstract}
\section{Introduction}

End-to-end autonomous driving learns planning directly from sensor data and has attracted sustained attention in both academia and industry~\cite{commaai_openpilot_2025,e2e_survey,hu2023_uniad,VAD}. Recent studies explore Vision-Language Model (VLM) driving agents \XRii{using visual question answering (VQA)} that combines the reasoning capability of large language models (LLMs) with visual perception 
to approximate human driving~\cite{GPT-4V-drive,VLM_drive_feedback_guided}. Chain-of-Thought
(CoT)~\cite{CoT} has been shown to enhance reasoning in LLMs~\cite{CoT_Explain}, and it is increasingly adopted in VLM driving agents to make the sequence of perception, analysis, and decision explicit~\cite{DriveLM,tian2024drivevlm,drivecot}. The intention is to strengthen planning 
\XRiii{by} improving controllability and interpretability. 
In this paradigm, the model 
first articulates a CoT 
reasoning \XRiii{for the current driving}, followed by the final planning trajectory. 
{Consequently, planning is taken for granted as causally driven 
through the preceding 
reasoning.}

Despite rapid progress, whether planning in current VLM driving agents is causally mediated by their reasoning remains insufficiently verified. {Existing works~\cite{DriveGPT4, Omnidrive}} primarily report trajectory quality and rule compliance, which assess how well the planning appears, but not which information pathway produce it. As a result, strong scores cannot be taken as evidence that reasoning contributes causally to planning.
Under these conditions, shortcut learning~\cite{shortcut_original,shortcut_LLM} is a central risk: a model can obtain high planning scores by exploiting biased or spurious priors, such as ego state and history, rather than by using reasoning
to construct the planning. 
\XRiii{Consequently, reasoning and planning may become decoupled, raising a critical safety concern: even when a plan appears safe, its underlying safety-relevant logic cannot be reliably verified, raising the risk that such outcomes are merely probabilistic coincidences rather than causally grounded decisions suitable for deployment.}

To \XRiii{systematically} investigate this causal link, a \XRii{high-quality VQA} dataset with plan-aligned CoT is necessary. 
However, existing datasets fall short of this need. Many real-world datasets~\cite{DriveLM, NuScenes-QA} use the nuScenes~\cite{nuScenes} benchmark as their foundation. While nuScenes provides high-fidelity sensor data, it inherently lacks the fine-grained semantic annotations, such as 
{traffic light states, } 
or complex lane topology, which are essential for deep reasoning. 
To circumvent the semantic limitations inherent to nuScenes, some researches~\cite{drivecot} have turned to simulation platforms such as CARLA~\cite{CARLA}. 
\XRiii{While highly controllable, these simulators suffer from a pronounced sim-to-real gap~\cite{CARLA-sim-to-real}: their trajectories follow idealized dynamics and fail to capture the nuanced 
nature of real-world human driving.}

\XRii{To bridge this gap, we introduce DriveMind, a high-quality dataset curated to facilitate the causal analysis of VLM-based driving agents. Built upon the nuPlan~\cite{nuPlan} benchmark that combines the authenticity of large-scale, real-world driving data with the rich, vectorized semantic context necessary for complex reasoning, DriveMind covers approximately $50$K samples spanning $61$ driving scenarios, providing broad and diverse coverage for analysis.} Another core contribution of DriveMind is the {modular}
organization of the dataset’s multi-modal inputs, structuring elements like visual data, ego state, and navigation into distinct 
{modules}. This design is critical for conducting controlled ablation studies. By selectively withholding specific information modalities, we can systematically dissect the information flow within a VLM agent and robustly attribute the final planning decisions to either high-level reasoning or low-level shortcut signals. 

Using our DriveMind dataset, we train and evaluate representative VLM-based driving agents~\cite{Qwen2.5-VL, llava, Omnidrive}. We uncover a striking result: an agent trained solely on textual priors, with no visual input and no CoT reasoning, achieves planning scores that match or even exceed those of a fully multimodal counterpart. This reliance on shortcuts is so entrenched that even applying an advanced policy alignment method {from reinforcement learning} (RL), Group Relative Policy Optimization (GRPO), fails to substantively restore the causal link from reasoning to planning.
These findings motivate our Reasoning-Planning Decoupling Hypothesis: the planning module {from an agent's output} predominantly relies on textual priors (i.e., ego state, history) as shortcuts, largely ignoring the visual context (i.e., surroundings, traffic signals) 
and the CoT reasoning.
To substantiate this hypothesis, we introduce a sequence-level attention analysis, which demonstrates the dominance of prior and ego-state tokens over visual and CoT tokens. 

Finally, we propose a generalizable, training-free diagnostic method to distinguish between genuine, CoT-grounded planning and shortcut learning, with the aim of establishing a plug-and-play standard for model evaluation.
Grounded in the principles of causal intervention, our method acts as a ``causal probe" by applying minor, {semantically plausible} perturbations to the textual priors (e.g., slight variations in ego states or historical positions). 
A planner that truly grounds decisions in visual evidence and CoT should be stable under such perturbations, whereas a shortcut-reliant planner will show brittle sensitivity to the exact prior pattern.
Therefore, a disproportionately large degradation in planning {scores} following perturbation indicates a high degree of shortcut learning. Conversely, a robust planning performance signifies that the agent's decisions are properly grounded in a holistic understanding of the driving scene. This method provides a simple yet powerful tool for assessing the causal fidelity of VLM agents. In summary, the main contributions of our paper are as follows:

\begin{itemize}[leftmargin=2.0em]
     \item We propose and validate the Reasoning-Planning Decoupling Hypothesis,
     \rev{which posits that current training paradigms are insufficient to forge a causal link between reasoning and planning, leading agents to instead learn shortcuts from textual priors.}
    
    \item We introduce DriveMind, a large-scale nuPlan-based dataset with plan-aligned  CoT  and a modular design, enabling causal analysis and systematic ablation  studies of VLM driving agents.

    \item We introduce the Causal Probe, a novel, training-free diagnostic method that detects shortcut reliance, providing a new tool for evaluating the causal robustness of driving agents..
\end{itemize}

\begin{figure*}[t] \centering 
\includegraphics[width=\textwidth,height=0.33\textwidth]{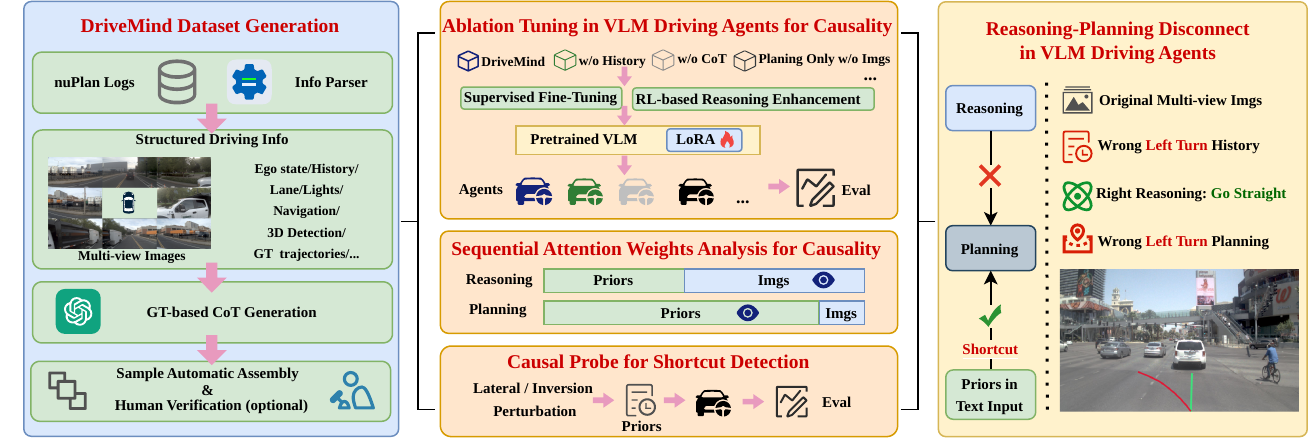}
    \caption{
    An overview of our framework for investigating the reasoning-planning disconnect in VLM agents. (Left) The DriveMind dataset generation pipeline. (Middle)  Our methodology for causal validation. (Right) An illustration of our core finding, where planning follows a textual prior shortcut, bypassing the correct CoT reasoning.
    }
    \label{fig:automatic_VQA_generation_pipeline}
\end{figure*}
\section{Related Work}
%
\paragraph{End-to-End Autonomous Driving.}
Unlike hierarchical driving models~\cite{ApolloAuto}, end-to-end autonomous driving aims to learn directly from sensor inputs and output planning trajectories, thereby avoiding error accumulation and facilitating optimization. Most existing end-to-end autonomous driving models are based on vector representation learning~\cite{e2e_survey}. 
Frameworks such as 
UniAD~\cite{hu2023_uniad} and VAD~\cite{VAD} transform perception information into Bird Eye View (BEV) feature vectors for subsequent trajectory planning.
ViDAR~\cite{ViDAR} and UAD~\cite{UAD} proposed self-supervised pre-training algorithms, enhancing the scalability of vector representation-based driving models.
More recently, the remarkable success of LLM~\cite{LLM_develop_1,LLM_develop_math,LLM_develop_code} and their multimodal extension, VLMs~\cite{vit,llava,vlm_survey_co_understand}, has introduced a new frontier. Researchers have begun to leverage the powerful feature extraction capabilities of VLMs to enhance the vectorized paradigm. 
For instance, models like VLP~\cite{VLP} and DiMA~\cite{DiMA}  distill prior knowledge from pre-trained VLMs into the driving model to improve planning performance. However, both the purely vectorized models and these early VLM-enhanced frameworks fundamentally rely on implicit feature transformations for decision-making. They lack an explicit, human-understandable reasoning process. 
\XRiii{Moreover, recent studies
reveal that such models are prone to shortcut learning on ego status~\cite{BEVPlanner} and spurious correlations~\cite{CaDeT,CausalVAD}. VLMs, 
generating explicit reasoning before planning, are thus regarded as 
promising 
for countering this by improving interpretability and controllability~\cite{AreVLMsReady}.}
\paragraph{Driving Agent with Reasoning.}
To enhance the interpretability of VLMs and adapt their capabilities for the driving domain, recent research has focused on 
optimizing models toward reasoning-based planning. Early approaches often employed multi-turn dialogue paradigms to elicit complex reasoning. For instance, DriveLM~\cite{DriveLM} introduced a graph-based VQA dataset that requires the model to first perceive and predict before a human manually selects key context for planning. Similarly, DriveVLM~\cite{tian2024drivevlm} proposed a hierarchical thinking dataset structured around scenario description, analysis, and planning. However, this multi-turn paradigm incurs significant data annotation and inference overhead, posing challenges to scalability and real-time application.
To improve efficiency, subsequent work has shifted towards single-turn dialogue incorporating CoT. DriveCoT~\cite{drivecot}, for example, explicitly integrates reasoning and planning into a single model response, but its reliance on the CARLA~\cite{CARLA} simulation environment raises key questions about sim-to-real transferability.
Omnidrive~\cite{Omnidrive} trains an agent for implicit reasoning via multi-task VQA. Meanwhile, Omnidrive recognizes the model's dependency on priors like ego state and attempts to mitigate this using counterfactual reasoning. However, as our subsequent experiments will reveal, this method has limited effectiveness in addressing shortcut learning from textual priors. Furthermore, its dataset is built upon nuScenes, inheriting the limitations of semantic richness required for deep causal analysis.
\section{Methodology}\label{sec:Methodology}
This section outlines our methodology, which includes {four} parts. First, we introduce the DriveMind dataset and its generation pipeline.  Second, we describe the {ablation} training of VLM driving agents on DriveMind. Third, we present our \XR{sequence-level} attention analysis for interpreting information flow. Finally, we introduce our perturbation-based causal probe for assessing shortcut learning.
\subsection{DriveMind Dataset Generation Pipeline}
Existing datasets either lack the complexity of the real world due to being simulation-based or lack the rich semantic information required to support deep reasoning, \XRiii{and they are not designed for causal analysis between reasoning and planning for VLM-based driving agents}. Therefore, the design of DriveMind adheres to four core principles, including a real-world foundation, rich semantic context, plan-aligned CoT, {plus} a 
{modular} structure for causal analysis.
As a large-scale planning benchmark released by Motional, nuPlan uniquely combines the high fidelity of real-world data with the rich semantics (e.g., traffic light status, lane topology)  necessary for deep reasoning. \XRiii{Based on nuPlan,} we convert the raw log data 
into the final samples of the DriveMind dataset through an automated data generation pipeline
, as illustrated in Figure~\ref{fig:automatic_VQA_generation_pipeline}. Our data generation pipeline consists of three main stages, beginning with scene parsing and feature extraction, followed by GPT-4.1-based CoT generation, and \XRiii{concluding with automatic sample construction with optional human verification.}


The first stage involves developing a parser to structure all key information from each nuPlan log. We extract ego-vehicle priors such as current and historical states, as well as the global navigation goal. For visual information, all camera images are first rescaled and then stitched into a three-row image grid corresponding to the front, side, and rear view groups to provide a spatially coherent input. 
\XRiii{For lane information, we encode it by identifying the ego lane and its neighboring lanes, along with their widths and directions relative to the vehicle heading, reflecting road topology and drivable regions.}
Additionally, we 
\XRiii{include }traffic signals by extracting the current state of relevant lights and the position of their corresponding stop lines.
\XRiii{Finally, we extract features of scene objects within a 20 m radius, categorizing valid targets into four groups: vehicles, pedestrians, bicycles, and other obstacles (e.g., traffic cones).}
These targets are localized using 3D annotations from nuPlan ground truth, and we then employ GPT-4.1 to identify visual characteristics (e.g., `a white SUV'
) of each object from their cropped images. This process enhances the agent's perception by creating an explicit link between each visual object and its corresponding textual description.


In the second stage, we use GPT-4.1 to \XRii{assist in generating} high-quality CoT. Recognizing that current LLMs face challenges in precise 3D spatial reasoning from images alone~\cite{VLM-weak-3D}, we ground the reasoning process by providing GPT-4.1 with structured scene ground truth. We design a comprehensive multimodal prompt that includes the preprocessed visual input, the structured scene ground truth from the first stage, and the expert's future planning trajectory. GPT-4.1 is tasked with explaining the causal logic behind the expert trajectory. The resulting CoT is organized into a structured text with three parts, which are scene ground truth, a causal analysis, and a macro decision.


Finally, we construct the final samples and perform human verification. To ensure the quality of the automatically generated CoT, a random 10\% ($\sim$5K) of the samples are manually reviewed by human experts, \XRiii{with only 2 out of ~5K reviewed samples (0.04\%) required regeneration},
confirming a high degree of logical correctness
of our \XRiii{generation pipeline}. 
The model is provided with two inputs, the visual inputs containing the preprocessed images and the textual priors containing ego state, history, and navigation information. 
The model is trained to produce two targets, 
the complete, human-verified ground truth CoT, and the ground truth planning
from the nuPlan human-expert driving logs.
The details of our reasoning generation process and an example including the complete input and target labels for a training sample are provided in Appendix~\ref{app:dataset_example}.


Our training set is constructed from $61$ driving scenarios sourced from the nuPlan training split. To mitigate the inherent long-tail distribution of this real-world data, we employ {square-root weighted stratified sampling}. For a target dataset size of $M=50,000$ across $K=61$ scenarios, the number of samples $n_i$ drawn from the $i$-th scenario (which originally contains $N_i$ samples) is calculated as:
\begin{equation}
\label{eq:sampling}
\textstyle n_i = M \cdot \frac{\sqrt{N_i}}{\sum_{j=1}^{K} \sqrt{N_j}},
\end{equation}
By this sampling, our DriveMind dataset maintains a high degree of complexity with 33.9\% hard scenarios (e.g., turnings and dense pedestrian interactions), 39.2\% general scenarios (e.g., slow down at lights), while simple scenarios (e.g., stationary) comprise only 26.9\%. This ensures that the model is trained primarily on scenarios requiring active reasoning rather than simple path following or being stationary. \XRii{Besides the dense semantic context derived from the real-world nuPlan logs and the high-quality ground truth based CoT, the critical feature of our dataset is its modular information structure, which is designed specifically for causal analysis. By explicitly separating different priors, reasoning, and planning within a coherent text, researchers can perform precise information ablation. This allows for a reliable determination of whether an agent’s planning stems from its autonomous visual perception and high-level reasoning, or from a reliance on ``shortcut learning" from the provided textual priors. The detailed distribution of all 61 scenario types and comparison with other datasets are provided in Appendix~\ref{app:dataset_proportion_details} and ~\ref{app:dataset_comparison}.}

\begin{table*}[t]
\centering
\scriptsize
\setlength{\tabcolsep}{11pt} 
\renewcommand{\arraystretch}{1.0} 
\caption{Overview of ablation settings for tuning analysis.  
\XRii{By systematically removing priors, 
ablation studies are conducted to
examine the impact of CoT and driving priors on VLM-based driving agents. The `decomposed CoT' means using multi sub-tasks for training.}
GRPO-based experiments confirm that the reasoning-planning disconnect effect is intrinsic rather than a superficial artifact.}
\label{tab:ablation_settings}
\begin{tabular}{@{}lccccc@{}} 
\toprule
\textbf{Agent} & \textbf{Base Model} & \textbf{Tuning} & \textbf{Vision} & \textbf{CoT} & \textbf{Priors} \\ 
\midrule
\multicolumn{6}{@{}l}{\textit{Supervised Fine-Tuning (SFT) Experiments}} \\
Base & Qwen2.5-vl-7b & --- & \cmark & --- & --- \\
CoT & Qwen2.5-vl-7b & SFT & \cmark & \cmark & All \\
Plan & Qwen2.5-vl-7b & SFT & \cmark & \xmark & All \\
Plan\_NoV & Qwen2.5-vl-7b & SFT & \xmark & \xmark & All \\
CoT\_NoHis & Qwen2.5-vl-7b & SFT & \cmark & \cmark & w/o History \\
CoT\_NoHis\_Ego & Qwen2.5-vl-7b & SFT & \cmark & \cmark & w/o His, Ego \\
CoT\_NoPri & Qwen2.5-vl-7b & SFT & \cmark & \cmark & None \\
CoT\_L & llava1.6-mistral-7b & SFT & \cmark & \cmark & All \\
Plan\_L\_NoV & llava1.6-mistral-7b & SFT & \xmark & \xmark & All \\
Omnidrive & Omni & SFT & \cmark & Decomposed & All \\
{Omnidrive\_NoPri} & Omni & SFT & \cmark & Decomposed  & None \\
\midrule
\multicolumn{6}{@{}l}{\textit{GRPO Policy Alignment Experiments}} \\
CoT\_grpo & Qwen2.5-vl-7b & SFT+GRPO & \cmark & \cmark & All \\
Base\_grpo & Qwen2.5-vl-7b & GRPO & \cmark & {Self-learned} & All \\
Base\_grpo\_NoPri & Qwen2.5-vl-7b & GRPO & \cmark & {Self-learned} & None \\
\bottomrule
\end{tabular}
\end{table*}
%
\subsection{Tuning VLMs for Causal Analysis}
Since general
VLMs are not pre-trained for  driving planning, a crucial first step is to fine-tune them into proficient driving agents.
Our primary objective, however, is to use the fine-tuning process itself as our object of study. We investigate how current training paradigms \XRiii{such as supervised fine-tuning (SFT) } shape the causal structure of an agent's learned behaviors, for testing our Reasoning-Planning Decoupling Hypothesis.
To this end, we conduct controlled experiments on a primary representative VLM, {Qwen2.5-vl}~\cite{Qwen2.5-VL}, and verify the universality of our findings on a second 
VLM, {Llava-1.6}~\cite{llava1.6}. Crucially, our setting is made comprehensive by also including {Omnidrive}~\cite{Omnidrive}, which employs {counterfactual reasoning} in an attempt to mitigate prior dependency. By analyzing it, 
we can test our hypothesis against existing state-of-the-art solutions. This three-pronged approach, which covers in-depth analysis, architectural generalization, and state-of-the-art comparison, provides a robust and sufficient foundation for \XRii{verifying} our claims.

 We first employ SFT to create a set of agents under various information ablation conditions.
 At this stage, we utilized the DriveMind dataset and its variants shown in Table~\ref{tab:ablation_settings}, where key input modalities, such as visual information, textual priors, or CoT are systematically removed. 
We fine-tune an agent with LoRA~\cite{LoRA} using each ablation dataset $D = \{((I_i, T_i), \mathcal{Y}_i)\}_{i=1}^{M}$, where $M$ is the total number of samples in the set. For each sample $i$, $I_i$ represents the visual input, $T_i$ is the textual prompt, and $\mathcal{Y}_i = (y_{i,1}, y_{i,2}, \dots, y_{i,L})$ is the corresponding target output sequence of length $L$.
With the pre-trained VLM parameters frozen as $\theta$ and the trainable matrices as $\phi$, we perform SFT, \XRii{and} the cross-entropy loss is used to compare two probability distributions for each output token:
\begin{equation}
\label{eq:sft_loss_explicit}
\textstyle \mathcal{L}_{\text{SFT}}(\phi) = - \frac{1}{M} \sum_{i=1}^{M} \sum_{j=1}^{L} \log P(y_{i,j} \mid I_i, T_i, y_{i,<j}; \theta, \phi).
\end{equation}

To test the robustness of this {disconnect} finding, we then incorporate an RL-based policy alignment stage using GRPO, which explicitly enhances an agent's reasoning capabilities by rewarding 
{the CoT process}. {We further sample $1,000$ new VQA samples from nuPlan, maintaining the same scenario distribution as DriveMind, and then do GRPO on both the SFT-trained model and the base model without SFT, with the latter serving as a baseline. Let $D_{grpo}$ denote the dataset and $\pi_\theta$ the policy, GRPO aims to maximize the following objective~\cite{deepseekr1_grpo}:
\begin{equation}
\label{eq:grpo_main}
\begin{split}
\textstyle \mathcal{J}_{\text{GRPO}}(\theta) = \mathbb{E}_{\substack{(I,T)\sim D\_grpo \\ \{o_i\}_{i=1}^{G}\sim\pi_{\theta_{\text{old}}}}} 
\bigg[ 
\frac{1}{G}\sum_{i=1}^{G}(\mathcal{A}_i\,\tilde{w}_i 
- \beta\,\mathbb{D}_{\mathrm{KL}}(\pi_\theta\|\pi_{\text{ref}}))
\bigg].
\end{split}
\end{equation}
This objective means that for each pair of inputs, $G$ outputs are sampled from an older policy $\pi_{\theta_{\text{old}}}$, then each output is evaluated its normalized advantage $\mathcal{A}_i$ that measures its quality relative to the group's average. The $\tilde{w}_i$ is a
weight to stabilizes updates by 
preventing overly large policy ratios and $\mathbb{D}_{\mathrm{KL}}$ is a KL-divergence penalty to regularize the current policy $\pi_\theta$ to stay close to the base policy $\pi_{\text{ref}}$ \XR{with the tunable hyperparameter $\beta$}. The model is encouraged to improve planning performance via enhanced reasoning by three rewards in $\mathcal{A}$: a location reward, a velocity reward, and a format reward. The first two rewards measure planning accuracy 
while the format reward incentivizes reasoning ability.
For detailed explanations of GRPO and our reward functions, please refer to Appendix~\ref{app:grpo_for_agents}.
}
This {stage} serves as a critical test to demonstrate that the observed disconnect is not an artifact of the SFT process, but a persistent characteristic that remains even after a direct intervention designed to strengthen the reasoning-planning link.


\subsection{Diagnosing with Sequence-level Attention Analysis}
To investigate our hypothesis and gain insight into the VLM’s internal mechanisms, we employ an interpretability framework called Sequence-level 
Attention Analysis. 
\XRii{It aims to analyze patterns of macro-level information flow between reasoning and planning in the agent’s output.}
When the model outputs trajectory tokens, we measure how much attention it pays to the preceding CoT tokens versus other contextual information (e.g., ego-state, history). A high degree of attention on the reasoning sequence would indicate a strong correlational link, suggesting the plan is at least 
conditioned on
the reasoning. Conversely, low attention would provide evidence in line with our decoupling hypothesis.

We formalize this by analyzing the attention weights of the VLM, which has $L$ transformer layers and $H$ attention heads per layer. Let $a^{(l,h)}_{j \to i}$ denote the raw attention weight that the token at position $i$ (the query) assigns to the token at position $j$ (the key) in layer $l$ and head $h$. Our analysis then proceeds in two steps.
First, we calculate a {\XR{sequence}-aggregated attention score}, $\bar{a}^{(l)}_{j \to S_t}$. This score represents the total attention that a target \XR{sequence} $S_t$ (e.g., the planning) directs to a single preceding token $j$, averaged over all $H$ attention heads in a given layer $l$:
\begin{equation}
\textstyle \bar{a}^{(l)}_{j \to S_t} = \frac{1}{H}\sum_{i\in S_t} \sum_{h=1}^H a^{(l,h)}_{j \to i}.
\end{equation}
%
Second, using this aggregated score, we compute our primary metric: the {proportional attention score}, $G^{(l)}_{S_M \to S_t}$. This value calculates the proportion of the target \XR{sequence's} ($S_t$) total attention that is allocated to a specific source \XR{sequence} $S_M$ (e.g., the {CoT reasoning}). Given a set of $M_{\mathrm{total}}$ preceding source \XR{sequences} $\{S_1, S_2, \dots, S_{M_{\mathrm{total}}}\}$, this attention proportion is given by:
\begin{equation}
\textstyle   G^{(l)}_{S_M \to S_t} = \left(\sum_{j \in S_M} \bar{a}^{(l)}_{j \to S_t}\right)/\left(\sum_{m=1}^{M_{\mathrm{total}}} \sum_{j \in S_m} \bar{a}^{(l)}_{j \to S_t}\right).
\end{equation}

This metric, which sums to 1 across all source \XR{sequences}, provides a direct and quantitative measure of information dependency. By comparing the attention proportion directed towards the reasoning \XR{sequence} versus other contextual \XR{sequences}, we can 
\XRiii{systematically} verify whether an agent’s planning is 
\XRiii{effectively} grounded in its own reasoning.

\subsection{Causal Probe for VLM Driving Agents}
\label{subsec:causal probe}
Standard performance metrics (e.g., planning scores) are insufficient to distinguish between planning that originates from genuine reasoning and that relies on shortcut learning. To efficiently and deeply probe an agent's decision-making logic and diagnose its degree of reliance on shortcuts, we introduce a novel, training-free {causal probe}. 
\XRii{The probe operates on the principle of robustness:} 
a planning decision 
\XRiii{effectively} driven by high-level reasoning from the visual scene 
should be largely invariant to minor, semantically plausible perturbations in its textual priors. Conversely, an excessive reaction to such minor changes indicates a brittle reliance on those priors.
Our framework introduces two distinct perturbation methods to test this principle.

The first method is a {quantitative test} we term \textit{lateral offset perturbation}. In this probe, we apply a small lateral offset, $\delta$, to the agent's ego-velocity. 
A robust agent, truly reasoning based on the visual scene and CoT, should be able to ignore or correct for a small offset in its ego velocity as its visual input clearly reflects the vehicle's true position and heading in the world. In contrast, an agent merely pattern-matching the textual priors would be misguided. 
Let $\mathbf{p}_i \in \mathbb{R}^2$ denote the predicted endpoint coordinates of the original input for the $i$-th sample, and $\tilde{\mathbf{p}}_i$ represent the endpoint under perturbation, we then quantify the impact by measuring the Mean Final Deviation (MFD):
\begin{equation}
\label{eq:mfd}
\textstyle
    \mathrm{MFD} = \frac{1}{n} \sum_{i=1}^{n} \lVert \mathbf{p}_i - \tilde{\mathbf{p}}_i \rVert_2,
\end{equation}
where $n$ is the number of tested samples. 
\XRii{We also evaluate the Relative Intervention Degree (RID) that is} defined as the ratio of the difference between a sample's output before and after perturbation to the theoretical error that would occur if the model perfectly followed the perturbation (i.e., $\delta \cdot v_i \cdot T$):
\begin{equation}
\textstyle
\mathrm{RID} = \frac{1}{n}\sum_{i=1}^{n} 
  \frac{\lVert \mathbf{p}_i - \tilde{\mathbf{p}}_i \rVert_2}{\delta \cdot v_i \cdot T} ,
\end{equation}
where $v_{i}$ is the ego-velocity of the $i$-th sample and $T$ is the prediction horizon. This normalized 
\XRii{RID quantitatively measures} the extent to which a model relies on textual priors, 
providing a scale-independent, robust metric that allows for cross-scenario comparisons.

The second method, \textit{lateral direction inversion}, is a {qualitative probe}. 
In this probe, we keep the ego-vehicle's current state (ego-state) unchanged but invert the lateral component of its historical trajectory 
\XRii{(e.g., mirroring a left-side merging history to a right-side one).}
A positive diagnosis for shortcut learning occurs when this inversion causes a directional change in the planning (e.g., 
a right turn 
), while the generated CoT that is primarily conditioned on the unchanged visual scene still presents correct arguments for a left turn. This exposes a stark contradiction between the agent's stated reasoning and its actual planning logic.

By combining the quantitative measurement and the qualitative diagnosis, we provide a systematic, training-free method to identify the presence and severity of shortcut learning in \XRii{VLM-based} driving agents. \XRiii{Together with our DriveMind dataset, they form a reusable evaluation framework.}

\section{Experiment Results and Analysis}\label{sec:Experiment}
%
\subsection{Experiment Setup}
\textbf{Implementation Details:} 
We fine-tune the agents listed in Table~\ref{tab:ablation_settings}, with the specific training parameters detailed in Appendix~\ref{app:training-details}. \XRii{We also verify our hypothesis on the large commercial model GPT-4o in a zero-shot manner.}

\textbf{Evaluation Protocol:} For evaluation, we use the official nuPlan Challenge test set. To ensure comprehensive coverage, we randomly sample $200$ scenarios for each of the $14$ distinct types defined in the benchmark. We conduct both open-loop and closed-loop (non-reactive) evaluations for every agent to holistically assess the performance. More details are provided in Appendix~\ref{app:nuplan_challenge}.
%
\subsection{Tuning Ablation Analysis Results}

\begin{table*}[t]
\centering
\scriptsize 
\setlength{\tabcolsep}{4pt} 
\renewcommand{\arraystretch}{1.05} 
\caption{Main ablation results on the nuPlan test set at 1s, 2s, and 3s horizons
}
\label{tab:all_results}
\begin{tabular}{@{}l|ccc|ccc|ccc|ccc@{}}
\toprule
& \multicolumn{3}{c}{\textbf{Open-Loop Score} $\uparrow$} & \multicolumn{3}{c}{\textbf{Closed-Loop Score} $\uparrow$} & \multicolumn{3}{c}{\textbf{Avg. ADE / FDE (m)} $\downarrow$} & \multicolumn{3}{c}{\textbf{Collision Ratio} $\downarrow$} \\
\cmidrule(lr){2-4} \cmidrule(lr){5-7} \cmidrule(lr){8-10} \cmidrule(lr){11-13}
\textbf{Driving Agent} & \textbf{1s} & \textbf{2s} & \textbf{3s} & \textbf{1s} & \textbf{2s} & \textbf{3s} & \textbf{1s} & \textbf{2s} & \textbf{3s} & \textbf{1s} & \textbf{2s} & \textbf{3s} \\
\midrule
Base & 36.52 & 33.62 & 31.50 & 59.80 & 42.48 & 36.99 & 2.67/4.45 & 4.80/8.81 & 7.02/13.75 & 7.84\% & 9.16\% & 9.87\% \\
CoT & 98.91 & {96.67} & {92.56} & {97.46} & {96.42} & {92.38} & {0.03/0.08} &{0.14/0.39} & {0.33/1.02} &{ 0.00\%} & {0.29\%} & {2.15\%} \\
Omnidrive & 98.92 & 96.85 & 92.96 & 97.49 & 96.35 & 92.47 & 0.03/0.08 & 0.13/0.38 & 0.31/0.99 & 0.00\% & 0.31\% & 2.03\% \\
\midrule
Plan & 98.89 & 96.64 & 92.58 & 97.38 & 96.49 & 92.60 & 0.04/0.09 & 0.14/0.41 & 0.33/1.02 & 0.00\% & 0.36\% & 2.55\% \\
{Plan\_NoV} & {98.95} & {96.84} & {92.92} & {97.46} & {96.37} & {92.54} & {0.03/0.08} & {0.13/0.38} & {0.32/0.98} & {0.00\%} & {0.36\%} & {2.45\%} \\
\midrule
CoT\_NoHis & 97.44 & 89.30 & 82.82 & 96.71 & 94.17 & 87.93 & 0.11/0.27 & 0.38/0.98 & 0.78/2.08 & 0.12\% & 1.62\% & 5.09\% \\
CoT\_His\_Ego & 74.51 & 61.65 & 57.70 & 90.39 & 87.23 & 81.32 & 0.69/1.25 & 1.35/2.60 & 2.04/4.05 & 0.24\% & 1.80\% & 5.27\% \\
{CoT\_NoPri} & {72.40} & {58.58} & {54.56} & {89.82} & {85.87} & {79.19} & {0.73/1.34} & {1.47/2.88} & {2.26/4.60} & {0.23\%} & {2.15\%} & {6.58\%} \\
{{Omnidrive\_NoPri}} & 75.91 & 61.34 & 57.32 & 89.99 & 86.52 & 79.69 & 0.64/1.17 & 1.29/2.53 & 2.00/4.14 & 0.31\% & 2.27\% & 6.54\% \\ 
\midrule
CoT\_L & 98.81 & 96.49 & 92.22 & 97.45 & 95.82 & 91.32 & 0.03/0.08 & 0.14/0.40 & 0.34/1.03 & 0.00\% & 0.54\% & 3.05\% \\
Plan\_L\_NoV & 98.91 & 96.90 & 93.03 & 97.54 & 96.53 & 92.87 & 0.03/0.08 & 0.13/0.36 & 0.31/0.95 & 0.00\% & 0.36\% & 2.09\% \\
\midrule\midrule
Base\_grpo & 96.38 & 85.31 & 77.84 & 96.74 & 92.52 & 84.51 & 0.15/0.35 & 0.47/1.20 & 0.95/2.50 & 0.12\% & 1.74\% & 6.22\% \\
{Base\_grpo\_NoP} & {39.93} & {29.70} & {27.84} & {73.88} & {61.75} & {55.44} & {1.80/3.22} & {3.55/6.88} & {5.39/10.68} & {10.59\%} & {15.68\%} & {20.41\%} \\
CoT\_grpo & 98.76 & 96.35 & 92.25 & 97.36 & 96.26 & 91.74 & 0.05/0.10 & 0.15/0.44 & 0.36/1.08 & 0.00\% & 0.48\% & 2.45\% \\
\bottomrule
\end{tabular}
\end{table*}

Our primary finding, detailed in Table~\ref{tab:all_results}, is the VLM agent's profound reliance on textual priors over CoT reasoning for planning. 
\XRii{The \textit{CoT} agent with explicit reasoning and \textit{OmniDrive}, trained using decomposed CoT, achieve performance nearly indistinguishable from the \textit{Plan} agent that produces no reasoning output.}
This comparison strongly indicates that the articulated reasoning process offers no discernible benefit to the final planning outcome. 
The most 
\XRiii{compelling} evidence for disconnect is a paradoxical result: the \textit{Plan\_NoV} agent, effectively `driving blind' without any visual input, performs on par with the  \textit{CoT} agent. This is a powerful demonstration of shortcut learning. Since the priors contain no information about environmental conditions, it reveals the model has learned to even bypass scene understanding by merely extrapolating from its current state and history. 
Conversely, the removal of textual priors results in a catastrophic performance collapse, as evidenced by the sharp decline across all metrics for \textit{CoT\_NoPri} and \textit{Omnidrive\_NoPri}. 
The consistency of these findings across diverse models, including Qwen-VL, Llava-1.6, and Omnidrive, provides robust evidence for our Reasoning-Planning Decoupling Hypothesis.

To further subject our hypothesis to a rigorous stress test, we employ GRPO, a powerful policy alignment tool. 
\XRii{By rewarding high-quality planning outcomes 
GRPO has the potential to optimize the 
model's} upstream cognitive chain, making it the most promising method for forging a causal link between reasoning and planning.
However, our experiment results directly contradict this optimistic outlook. While the \textit{Base\_grpo} agent shows superficial improvements over the \textit{Base} counterpart, this proves to be a ``false prosperity" built on shortcuts. Once textual priors are removed, the \textit{Base\_grpo\_NoPri} agent's performance collapses catastrophically, falling well below even the original \textit{Base} model. This decisively shows that this powerful alignment tool does not guide the model to learn deeper visual reasoning; instead, it rewards 
model's reliance on textual shortcuts, exacerbating the overfitting problem. Furthermore, the \textit{CoT\_grpo} agent's planning capability is even weaker than that of \textit{CoT}, 
suggesting that 
\XRii{GRPO is unable to effectively link reasoning and planning to search better results}.
Ultimately,  
\XRii{this stress test provides strong evidence that the reasoning–planning disconnect is a fundamental characteristic of the current VLM fine-tuning paradigm.}
Even advanced RL policy 
designed to 
boost alignment fails to rectify this disconnect and may even intensify the model's dependence on shortcuts. 
\XRiii{Appendix~\ref{app:extended_other_exps} provides additional results supporting our findings.}

\begin{table*}[t]
    \centering
    \scriptsize
    \renewcommand{\arraystretch}{1.0}
    \caption{
    Planning performance comparison under visual inputs and reasoning perturbations.}
    \label{tab:visuai_cot_perturbations}
    \begin{tabular}{lcc}
        \toprule
        \textbf{Agent} & \textbf{Open-Loop Score (3s)} $\uparrow$ & \textbf{Close-Loop Score (3s)} $\uparrow$ \\
        \midrule
        CoT & 92.56 & 92.38 \\
        CoT\_V\_Crop\_Noise & 91.86 & 90.79 \\
        CoT\_V\_Replace & 92.36 & 92.27 \\
        CoT\_R\_Replace & 92.39 & 92.17 \\
        \bottomrule
    \end{tabular}
\end{table*}
\subsection{Perturbation to Visual Inputs and Generated Reasoning}
\label{subsec:visuai_cot_perturbations}
To further validate our findings regarding the reasoning-planning disconnect, we conduct intervention experiments by perturbing both the visual inputs and the intermediate reasoning traces during inference, evaluating whether the final planning genuinely relies on visual scene understanding or generated reasoning. Specifically, we evaluate three perturbation variants on the \textit{CoT} agent: (1) \textit{CoT\_V\_Crop\_Noise}, which randomly crops the input images and injects Gaussian noise; (2) \textit{CoT\_V\_Replace}, which randomly substitutes the original visual scene with a contradictory one; and (3) \textit{CoT\_R\_Replace}, where we replace the macro-driving decision within the generated reasoning trace with an incompatible decision taken from a different driving scenario while keeping the visual input intact, and then allow the agent to continue generating the final planning.

Table~\ref{tab:visuai_cot_perturbations} reports the planning performance across all perturbation settings. Under visual perturbations, the \textit{CoT} agent exhibits negligible performance changes in both open-loop and closed-loop scores. Together with the findings from \textit{Plan\_NoV} that completely strips visual input and serves as the most extreme visual perturbation, these results collectively demonstrate that the planning ability of the trained VLM driving agent is effectively visually inert. Moreover, despite replacing the generated reasoning with a contradictory macro-decision, \textit{CoT\_R\_Replace} maintains high performance with 92.39 and 92.17 in open-loop and closed-loop scores respectively, suggesting that the final planning process operates largely independently of the generated reasoning. Together, these results provide further support for our Reasoning-Planning Decoupling Hypothesis.

\subsection{Sequence-level Attention Analysis Results}
\begin{table*}[t]
\centering
\scriptsize 
\renewcommand{\arraystretch}{1.0} 
\setlength{\tabcolsep}{2pt} 
\caption{
    Attention distributions across preceding sequences during target sequence generation.
}
\label{tab:attn_cot}
\begin{tabular}{@{}lcccccc@{}}
\toprule
\multirow{2}{*}{\textbf{Attention Target}} & \multicolumn{3}{c}{\textbf{Scenario Reasoning Task}} & \multicolumn{3}{c}{\textbf{Planning Task}} \\
\cmidrule(lr){2-4} \cmidrule(lr){5-7}
& Shallow Layer& Middle Layer& Final Layer& Shallow Layer& Middle Layer& Final Layer \\
\midrule
Image Tokens (\%) & \textbf{2.91} & \textbf{5.55} & \textbf{11.52} & 2.47 & 2.51 & 1.82 \\
Textual Priors (\%) & 3.74 & 5.34 & 10.36 & \textbf{11.52} & \textbf{17.93} & \textbf{26.97} \\
Generated Reasoning (\%) & ---&---&--- & 17.11 & 21.12 & 19.37 \\
All Textual Tokens (\%) & 97.09 & 94.45 & 88.48 & 97.53 & 97.49 & 98.18 \\
\bottomrule
\end{tabular}
\end{table*}
To investigate the agent's decision-making process, we analyze the attention flow of the \textit{CoT\_grpo} agent. We partition the model's output into two key \XR{sequences}: the \textit{Reasoning} (CoT) \XR{sequence} and the \textit{Planning} (trajectory) \XR{sequence}. We then measure the proportion of attention each \XR{sequence} directs towards different preceding information sources (Images, Textual Priors, etc.) at shallow, middle, and final layers of the model, allowing us to observe how information dependencies evolve.

As shown in Table~\ref{tab:attn_cot}, when the agent generates the \textit{Reasoning} \XR{sequence}, its attention to the {Image Tokens} progressively increases, from $2.91\%$ in the shallow layer to over $11\%$ in the final layer. This indicates that the reasoning process is continuously and increasingly grounded in the visual input. While attention to {Textual Priors} also grows, it remains secondary to the visual information in deeper layers. This behavior represents a healthy, logical process where the agent builds its reasoning primarily upon what it sees, while integrating priors as necessary context.

In contrast, the attention shifts dramatically during the \textit{Planning} phase. Attention to the model's generated {Reasoning} first increases and then fades, while attention to the {Textual Priors} skyrockets from $11.52\%$ to nearly $27\%$ in the final layer. This shows the model discovering the decisive role of priors and  becoming highly dependent on them. Simultaneously, attention to the {Image Tokens} becomes negligible, dropping below $2\%$, which indicates that the model barely requires visual information for planning. This finding, 
with the high performance of the \textit{Plan\_NoV} agent in Table~\ref{tab:all_results}, leads to a 
\XRiii{clear} conclusion: the VLM driving agent relies almost exclusively on textual priors for planning, rendering the reasoning and planning processes almost entirely disconnected.

A consistent observation across both tasks is the VLM's overwhelming preference for attending to textual tokens over image tokens. This highlights an inherent {modality bias} in current VLMs. This insight suggests that a potential path to mitigating shortcut learning in VLM driving agents is to 
force the visual understanding capabilities \XRiii{especially when planning}.
\begin{figure*}[t] \centering
\includegraphics[width=\textwidth,height=0.14\textwidth]{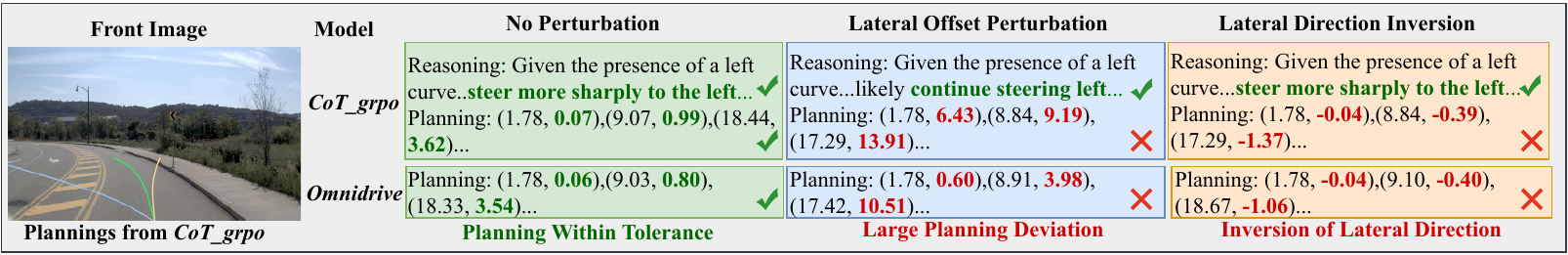}
    \caption{
   Comparison of an agent’s original plan and its trajectories under two perturbations.
   }
    \label{fig:calsual_prob}
\end{figure*}
\subsection{Causal Probe Detection Results}
To evaluate the diagnostic efficacy of our causal probe,
we select two key agents for this analysis: the strongest reasoning agent, \textit{CoT\_grpo}, and a state-of-the-art baseline known to include measures against prior dependency, \textit{Omnidrive}.
First, we apply the \textit{lateral offset perturbation}. In this test, we add a small lateral velocity, $\delta$, perpendicular to the vehicle's heading, scaled to the ego-vehicle's current speed $v_{\text{ego}}$ (i.e., $\delta = 0.1 \cdot v_{\text{ego}}$). 
This creates a small and semantically plausible perturbation. As shown in Figure~\ref{fig:calsual_prob}, both agents exhibit extreme sensitivity to this dynamic perturbation. 

\rebuttal{For more precise quantitative analysis and to test the model's sensitivity under different perturbation levels, we test on the \textit{CoT\_grpo} utilizing a test set of 300 randomly sampled scenarios with varied the perturbation magnitude $\delta$ from 0.02 to 0.3, and we report the MFD and RID in Table~\ref{tab:delta-offset—perturbation}. From the results, the model exhibits extreme instability even at low perturbation levels. As $\delta$ increases from 0.02 to 0.06, the  MFD jumps significantly to 2.17 m. This confirms that the agent is captured by the textual prior almost immediately, lacking the 
robustness to correct for even subtle semantic conflicts.  As $\delta$ increases further 
the MFD continues to grow monotonically, reaching 15.46m at $\delta=0.3$. 
The RID initially peaks at 2.17 for $\delta=0.1$, then plateaus or slightly decreases to 2.04 for $\delta=0.3$. Actually, this plateau does not indicate robustness as the RID is still larger than 1; rather, it reflects saturation of “wrong outputs” as the agent predicts extreme trajectories beyond its learned manifold, lagging behind increasing perturbations.}
The disproportionate reactions of the driving agents under small lateral offset perturbation from both the qualitative and quantitative results strongly indicate the agents' heavy reliance on the ego states in text priors for shortcut learning.

Next, we apply the \textit{lateral direction inversion} probe. After inverting the lateral component of the historical trajectory, we observed that both \textit{CoT\_grpo} and \textit{Omnidrive} consistently reversed their planned maneuver direction.
Most critically, a manual inspection of the \textit{CoT\_grpo} agent's outputs after this inversion revealed a direct contradiction \rebuttal{in 100\% of the tested samples}. In these cases, the agent would generate a correct CoT based on the visual scene (e.g., reasoning for a left turn) but then produce a plan that incorrectly followed the perturbed history (e.g., executing a right turn). 

This finding provides 
\XRiii{compelling} evidence for our {Reasoning-Planning Decoupling Hypothesis}. It not only demonstrates the deep-rooted nature of the problem but also validates the efficacy of our causal probes in uncovering these critical failures in reasoning. More examples are shown in Appendix~\ref{app:examples-of-causal-probe}.

\begin{table*}[t]
    \centering
    \scriptsize
    \setlength{\tabcolsep}{10pt}
     \caption{Impact of different perturbation intensities ($\delta$) on \textit{CoT\_grpo}. The large MFD and RID demonstrate its strong reliance on text priors.}
    \label{tab:delta-offset—perturbation}
    \begin{tabular}{cccccc}
        \toprule
$\delta$        & \textbf{0.02} & \textbf{0.06} & \textbf{0.08} & \textbf{0.10} & \textbf{0.30} \\
        \midrule
        \textbf{MFD} $\downarrow$ & 0.55 & 2.17 & 3.75 & 5.89 & 15.46 \\
        \textbf{RID} $\downarrow$ & 1.00 & 1.47 & 2.04 & 2.17 & 2.04 \\
        \bottomrule
    \end{tabular}
\end{table*}

\subsection{\rebuttal{Evaluation on Larger Models}}
\rebuttal{To assess whether our findings generalize to larger VLMs, we apply the same causal probes to the substantially larger GPT-4o in a zero-shot setting. Under perturbation, GPT-4o produces correct scenario reasoning (e.g., “The ego-vehicle is in a lane moving straight”), yet its predicted trajectory still adhered to the falsified textual prior, indicating that the reasoning–planning disconnect persists beyond the 7B model scale. Quantitatively, with a perturbation level of $\delta=0.1$, GPT-4o yields an MFD of 2.55 m and an RID of 1.35. This zero-shot evaluation provides a necessary assessment of a closed-source model, demonstrating that the disconnect is fundamental to both open-source and proprietary large-scale VLM architectures.}
\section{Conclusion}
\label{sec:conclusion}

\XRiii{In this paper, we have investigated the causal link between reasoning and planning in VLM driving agents. To enable this analysis, we introduce DriveMind, a benchmark designed for causal analysis, and develop diagnostic tools including controlled ablations and perturbation-based probes. Our experiments provide strong evidence for the {Reasoning-Planning Decoupling Hypothesis}: agents learn to rely on textual priors as shortcuts for planning, while their generated CoT serves as a plausible but non-causal byproduct. This work reveals that perceived interpretability of current agents can be misleading, underscoring the need for causally-aware training paradigms to build truly robust VLM driving agents. Possible future directions and limitations are discussed in Appendix~\ref{sec:future_work} and ~\ref{app:limitations}.
}
\section*{Acknowledgments}
This research is supported by cash and in-kind funding from NTU S-Lab and industry partner(s). 
The authors declare no competing interests.

\bibliographystyle{plainnat}

{\small
\bibliography{citations}
}

\appendix
\section{Appendix}

\subsection{\XRii{Future Work}}
\label{sec:future_work}
Our future work will focus on resolving the reasoning-planning disconnect. We plan to pursue two research directions aimed at developing a more robust and causally-faithful VLM driving agent.

\textbf{Mitigating Modality Bias via Contrastive Pre-Finetuning.}
Our main results revealed a key finding: an agent with no visual input (\textit{Plan\_NoV}) can achieve planning scores nearly identical to a fully multimodal agent. 
This strongly suggests that the models have learned to almost entirely disregard the visual modality. Therefore, a critical future direction is to introduce a dedicated {contrastive pre-finetuning stage} designed to make the visual input indispensable. In this stage, we will train the VLM on a curated, driving-related dataset where the textual input is held constant while the visual input varies to produce a different correct output. This process will force the model to extract decisive information from the visual modality, rebalancing its modality dependence and enhancing the salience of the vision encoder before the downstream driving SFT/GRPO begins.

\textbf{Breaking Shortcuts via Contrastive Learning.}
Our second direction is to enhance the agent's SFT/GRPO process via {contrastive learning}, leveraging negative examples to guide its policy. This involves augmenting the training data with {verified conflict samples}. For a target scene (e.g., {Image\_A}), we pair it with a mismatched textual prior (e.g., {Priors\_B}) from another scenario whose original ground-truth trajectory ({Plan\_B}) has first been verified via the nuPlan metrics as an unsafe plan for the target scene. For these augmented samples, we will implement a {repulsive loss function}. This objective penalizes the policy if its predicted trajectory falls within a certain distance margin of the undesirable trajectory, {Plan\_B}. 
The agent is forced to abandon the ``prior equals plan'' shortcut and instead learn to genuinely reason, aiming to connect the {reasoning and planning}.

Furthermore, these two directions are highly complementary. Mitigating modality bias aims to improve the information flowing into the reasoning module, while the contrastive learning strategy aims to enforce that the plan is correctly derived from it. Their combination promises a driving agent whose planning is grounded in its own reasoning.

\XRiii{
\subsection{Limitation}\label{app:limitations}
Our primary fine-tuning experiments employ 7B-parameter models and Omnidrive. Fine-tuning of larger models (e.g., 32B+) remains to be explored. Our diagnostic framework is model-agnostic and can be directly applied to evaluate such models as they become available. Notably, the disconnect already manifests in GPT-4o under zero-shot evaluation, suggesting that model scale alone does not resolve the issue.
}

\XRiii{
\subsection{Discussion}\label{app:discussion}
\paragraph{The high performance score on benchmarks is not enough.}
Existing planning metrics mainly evaluate the overlap between predicted and expert trajectories, rather than whether the predicted plan is actually causally grounded in scene understanding. For example, such metrics cannot distinguish genuine reasoning-based planning from shortcut learning. In benchmark settings, this creates a systematic gap from real driving: priors are often strongly correlated with expert trajectories, environmental changes are limited, and performance is judged mainly from final trajectory similarity. Under such conditions, a model can achieve high scores by exploiting priors or memorized trajectory patterns, even without truly understanding the scene.

Our experiments provide direct evidence for this. Even the blind \textit{Plan\_NoV} agent, which has no visual input, performs comparably to the full \textit{CoT} agent, while removing priors causes substantial degradation. Moreover, in the Scene Replacement experiment in~\ref{subsec:visuai_cot_perturbations}, agents still achieve strong planning scores even when the textual priors are paired with completely mismatched visual scenes, where those scores should have dropped sharply if planning were truly scene-grounded. These results show that the apparent `good performance' is superficial: it reflects weakness of the current evaluation environment rather than genuine safe planning.

\paragraph{The importance of diagnosing the disconnect.}
Reasoning and planning do need to be aligned for safety in VLM driving models. In autonomous driving carrying human lives, safety is not only about obtaining a good trajectory in a benchmark; it is life-critical, so being able to verify that the trajectory is produced for the right human-logic reasoning is essential. If reasoning and planning are disconnected, then humans cannot reliably verify whether the vehicle is safe under the logic by which safety should be judged. In that case, an seemingly safe plan may only be a probabilistic coincidence, rather than a safety-guaranteed decision that can be trusted for deployment.

Therefore, the reasoning-planning disconnect is a critical issue precisely because it shows that current "good planning performance" does not come from genuine scene understanding; such performance cannot be trusted as evidence of safe planning in real-world driving. This is why we design the causal probe in this paper: a model should be considered truly strong only if it both achieves high planning scores and passes causal verification showing that its planning is genuinely aligned with scene-grounded reasoning.

\paragraph{Comparison with Other Related Work.} Concurrent efforts such as Drive-R1~\cite{Drive-R1} and Alpamayo-R1~\cite{Alpamayo-R1} report similar observations to ours and demonstrate improvements in planning performance (e.g., L2 distance and collision rate). These works primarily focus on enhancing model capability, while our work takes a complementary perspective by investigating the underlying reasoning–planning disconnect. In particular, prior studies evaluate improvements through standard planning metrics, but do not explicitly examine whether planning is causally grounded in scene understanding via perturbation or interventional analysis. In contrast, we formally define the reasoning–planning disconnect in VLM driving agents and provide a systematic diagnostic framework to measure it. Our causal probes serve as a validation mechanism to assess whether performance gains are supported by faithful reasoning, offering a foundation that can complement and strengthen future solution-oriented approaches.
}

\subsection{Example data of DriveMind Dataset and Generation Details}\label{app:dataset_example}
This section offers a detailed overview of the DriveMind dataset, covering both its structure and its generation process.

\begin{figure*}[t] \centering
\includegraphics[height=0.99\textheight]{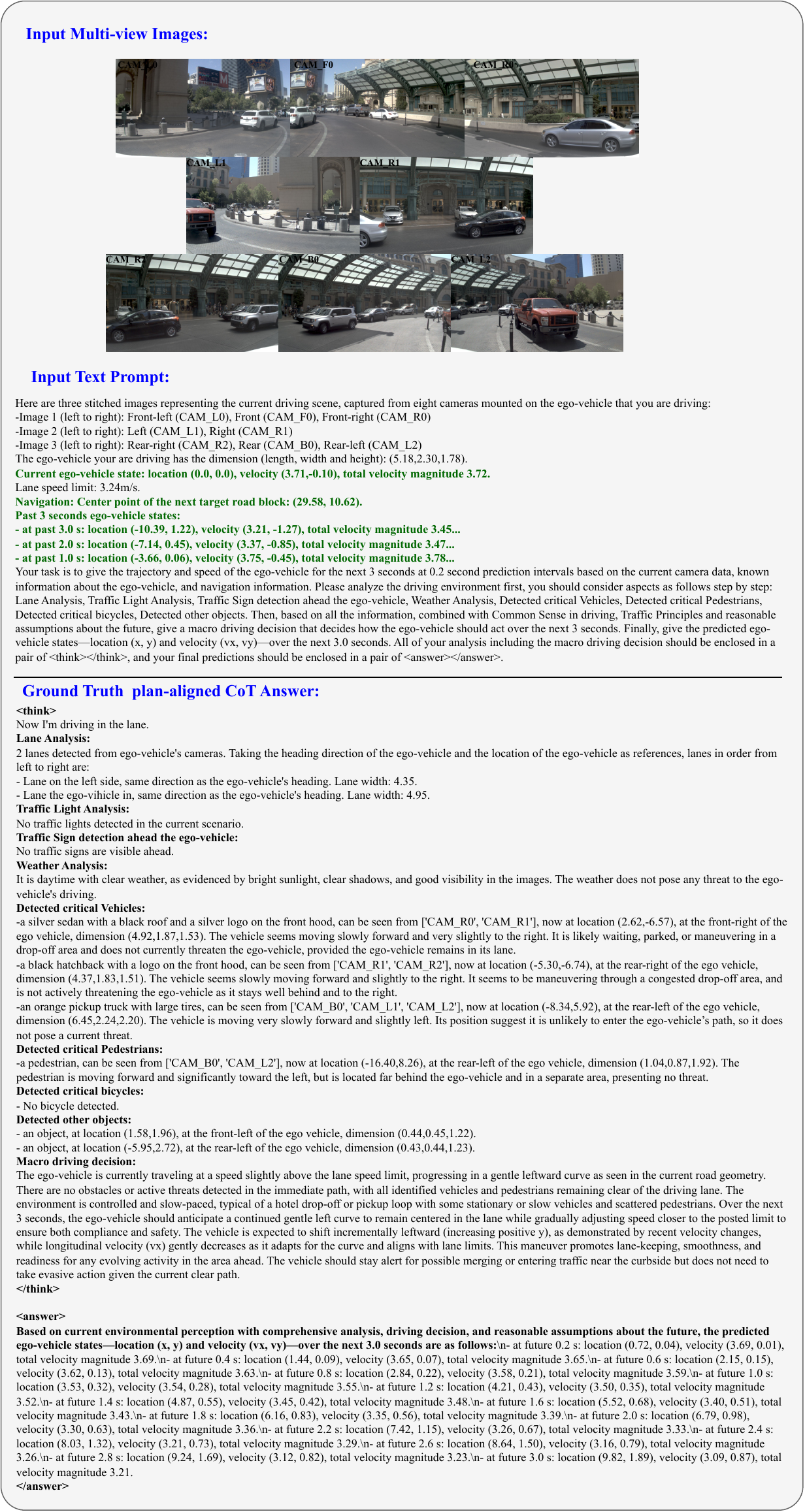}
    \caption{An example of our plan-aligned CoT driving VQA data from DriveMind, with the three priors in the text input marked as green.}
    \label{fig:example_data}
\end{figure*}

Figure~\ref{fig:example_data} illustrates an example of data from the DriveMind dataset. Our input consists of concatenated surround views and textual inputs (including prior knowledge and instructions). 
The target output is structured into two parts. First, a plan-aligned CoT is presented within a \texttt{<think>} block. This CoT incorporates a comprehensive analysis of the scene, including key static elements (e.g., lanes, traffic lights, road signs), environmental conditions (e.g., weather), and critical dynamic road users (e.g., nearby vehicles, pedestrians, and bicycles), concluding with a macro-level driving decision. Second, following this reasoning, the final trajectory plan is provided within an \texttt{<answer>} block.
All the reasoning and planning are causally linked through the macro driving decision and the explicit causal logic-enhancing sentence. We highlight in green the information discussed in the main text that may contribute to shortcut learning: ego state, historical data, and navigation information. Our ablation experiments in the main text confirm that VLM driving agents' planning indeed originates from shortcut learning of textual priors, with reasoning emerging as a logical byproduct.

\begin{figure*}[t] \centering
\includegraphics[width=\textwidth]{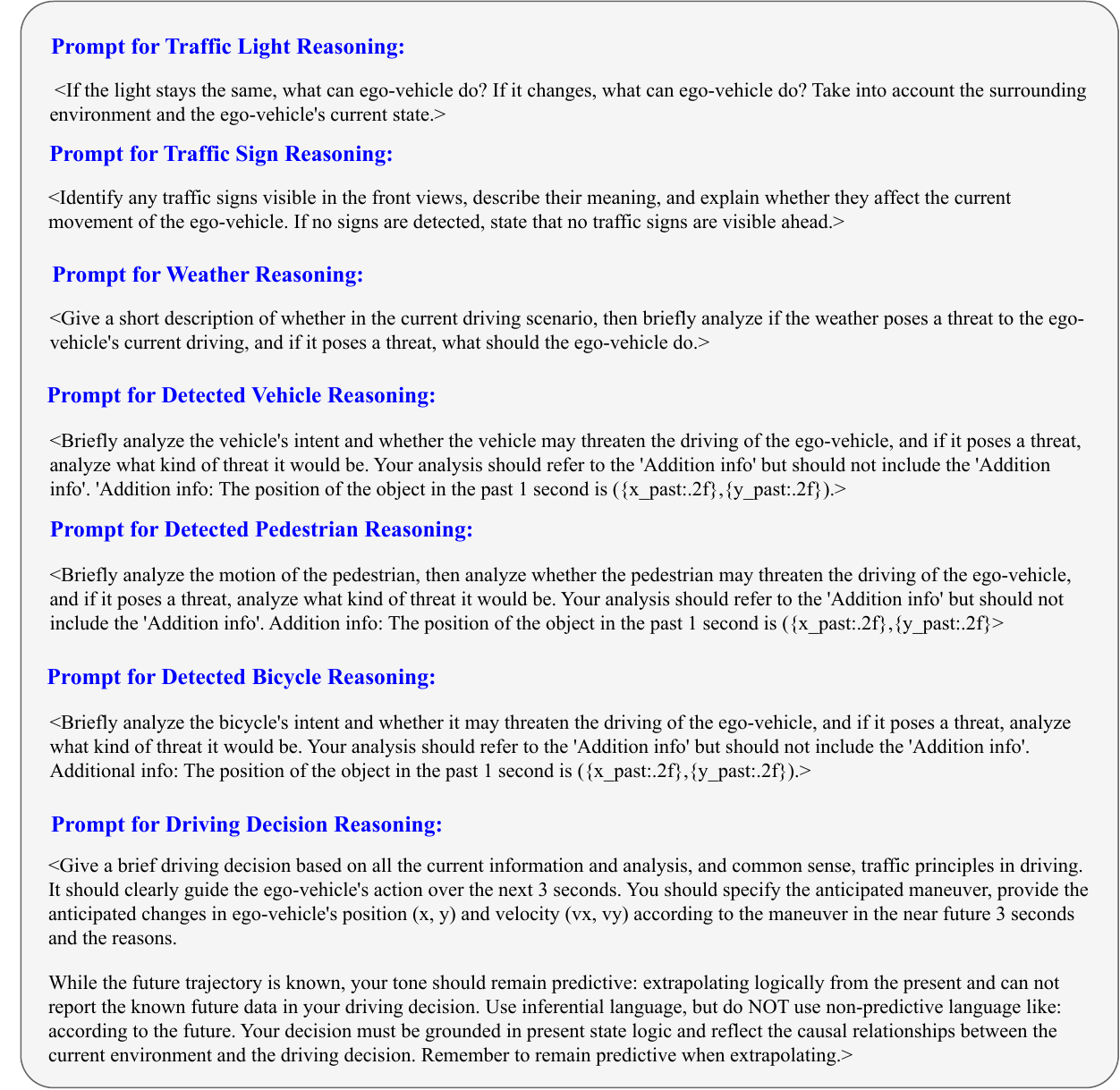}
    \caption{Prompt for GPT-4.1 to generate analysis for different parts. All together combine to form the plan-aligned CoT in DriveMind.}
    \label{fig:prompt_for_reasoning}
\end{figure*}

To generate the CoT data, we first use the info parser to extract and make the structured driving info from the nuPlan ground truth log. Then, we use the prompts illustrated in Figure~\ref{fig:prompt_for_reasoning} to instruct the GPT-4.1 to generate scenario-aware analysis of the target part, such as weather and detected vehicles. We input the images and corresponding ground truth from the structured info for each part to address the visual-spatial perception limitations inherent in GPT-4.1, and let it generate analyses for each part in sequence. When generating the macro driving decision, all preceding environmental ground truth perceptions, analyses, and the future trajectory are incorporated as inputs, with the given future trajectory ensuring the accuracy of the decision. Notably, to ensure the causality of the reasoning towards driving, we imposed the following constraints on GPT-4.1 when letting it give the macro driving decision: `While the future trajectory is known, your tone should remain predictive: extrapolating logically from the present and can not report the known future data in your driving decision. Use inferential language, but do NOT use non-predictive language like: according to the future. Your decision must be grounded in present state logic and reflect the causal relationships between the current environment and the driving decision. Remember to remain predictive when extrapolating.' These operations ensure both the accuracy and the causality of our macro driving decision. After all the reasoning generated, we combine them with the ground truth information, plus a causal logic-enhancing sentence, to form the CoT data. Finally, a portion of the data will undergo rigorous manual verification to ensure content and logical accuracy.

\XRiii{
For the human verification, we randomly sampled ~5K samples (10\% of the 50K dataset) for human review. Annotators with driving experience assessed each sample on three criteria: (a). accuracy of scene entity descriptions against the visual input, (b). logical correctness of the CoT reasoning chain, and (c). consistency between the reasoning and the ground-truth expert trajectory. Samples failing any criterion were flagged and regenerated. Only 2 out of ~5K reviewed samples (0.04\%) required regeneration. The extremely low regeneration rate confirms the pipeline produces reliable output at scale, meaning human review can be safely omitted during large-scale expansion.

A key design choice explains the high quality: our pipeline first extracts structured ground truth from nuPlan logs such as 3D object positions, lane topology, traffic light states, etc., and provides all of this to GPT-4.1. With these factual scene elements already established, GPT-4.1's task is narrowly scoped: it only needs to identify the most important causal factor linking these known elements to the given expert trajectory and articulate it as a reasoning chain. This highly constrained generation space leaves minimal room for hallucination or factual error.
}


\subsection{Details of the Scenario Distribution of DriveMind Dataset}\label{app:dataset_proportion_details}
\rebuttal{
We report the exact number of samples associated with each scenario category in the DriveMind dataset, organized by difficulty level:
\begin{itemize}[leftmargin=2em]
    \item Hard: High-dynamic scenarios involving complex interactions (e.g., turning, lane changing).
    \item General: Standard driving scenarios governed by clear traffic rules (e.g., following lane with lead, stopping at lights).
    \item Simple: Low-dynamic or static scenarios (e.g., stationary in traffic, low magnitude speed).
\end{itemize}
The corresponding proportions of scenarios across these difficulty levels are also provided. Our square-root–weighted stratified sampling strategy (see Eq.~(\ref{eq:sampling})) yields a well-balanced distribution across difficulty levels, resulting in 33.9\% Hard, 39.2\% General, and 26.9\% Simple scenarios. Detailed statistics are presented in Table~\ref{tab:scenario_distribution}. 
}

\begin{table*}[t]
    \centering
    \scriptsize
    \caption{\rebuttal{Comparison of datasets in terms of real-world logs, semantic richness, expert CoT availability, modularity, and scenario coverage.}}
    \label{tab:dataset_comparison}
    \resizebox{\textwidth}{!}{
    \begin{tabular}{l c c c c c} 
        \toprule
        \textbf{Dataset} & 
        \textbf{Real-World Logs} & 
        \textbf{Semantic Richness} & 
        \textbf{Expert CoT} & 
        \textbf{Modular Structure} &
        \textbf{Scenario Coverage} \\
        \midrule
        DriveLM        & Mixed & Moderate (nuScenes), High (CARLA) & Yes & No & --- \\
        DriveVLM       & Yes   & Moderate (SUP-AD)                 & Yes & No & 40  \\
        DriveCoT       & No    & High (CARLA)                      & Yes & No & 5   \\
        OmniDrive      & Yes   & Moderate (nuScenes)               & No  & No & --- \\
        \textbf{DriveMind (Ours)} & \textbf{Yes} & \textbf{High (nuPlan)} & \textbf{Yes} & \textbf{Yes} & \textbf{61} \\
        \bottomrule
    \end{tabular}
    }
\end{table*}

\subsection{\XRii{Comparison with Other Datasets}}\label{app:dataset_comparison}
\XRii{Table ~\ref{tab:dataset_comparison} provides a detailed comparison between DriveMind
and other VQA datasets in terms of real-world logs, semantic richness, expert CoT availability, modularity, and scenario coverage. DriveMind distinguishes itself from existing datasets through several characteristics:

\textbf{Real-world fidelity with semantic richness:} Unlike simulation-based datasets (e.g., DriveCoT), DriveMind is built upon nuPlan's real-world logs and utilizes its ground-truth annotations with high semantic richness. This not only enriches the environmental semantic information contained within DriveMind, such as stopping line locations and traffic light status, but also ensures the absolute accuracy of the information, both of which significantly ensures the high-quality of DriveMind's VQA.

\textbf{More diverse scene categories:} DriveMind covers 61 distinct driving scenarios, significantly surpassing other datasets such as DriveVLM (40) and DriveCoT (5). Combined with our sampling mechanism, DriveMind incorporates more effective long-tail scenarios and complex scenarios, so that it can provide a more robust foundation for training and evaluating for VLM-based driving agents. 

\textbf{Unique modular structure:} DriveMind employs a unique modular design that integrates diverse textual priors, reasoning processes, and planning processes into semantically coherent dialogues. This enables researchers to perform precise information ablation and conduct the causal probe experiments to evaluate a VLM driving model.

\textbf{High-quality CoT with human verification:} DriveMind provides expert CoT annotations generated with assistance from the latest multi-modal LLM GPT-4.1, whereas prior datasets typically rely on older models such as GPT-3.5. In addition, our samples also undergo manual verification, ensuring that the expert CoT annotations are of substantially higher quality.
}

\begin{table*}[p] 
\centering
\caption{\rebuttal{Distribution of scenario types and difficulty-level proportions in the DriveMind.}} 
\label{tab:scenario_distribution}
\tiny
\resizebox{0.87\textwidth}{!}{%
\begin{tabular}{cccc}
\hline
Difficulty Category & Scenario Name & Sample Count & Total/Proportion \\ \hline
 & changing\_lane\_to\_left & 56 &  \\
 & changing\_lane\_to\_right & 51 &  \\
 & changing\_lane\_with\_trail & 24 &  \\
 & crossed\_by\_vehicle & 41 &  \\
 & high\_lateral\_acceleration & 648 &  \\
 & starting\_protected\_cross\_turn & 291 &  \\
 & starting\_protected\_noncross\_turn & 434 &  \\
 & high\_magnitude\_jerk & 33 &  \\
 & near\_barrier\_on\_driveable & 274 &  \\
 & near\_construction\_zone\_sign & 1033 &  \\
 & near\_long\_vehicle & 1739 &  \\
 & near\_multiple\_pedestrians & 395 &  \\
 & near\_pedestrian\_at\_pickup\_dropoff & 1474 &  \\
 & near\_pedestrian\_on\_crosswalk & 2619 &  \\
 & near\_pedestrian\_on\_crosswalk\_with\_ego & 267 &  \\
 & near\_trafficcone\_on\_driveable & 1251 &  \\
 & starting\_high\_speed\_turn & 206 &  \\
 & starting\_unprotected\_cross\_turn & 337 &  \\
 & starting\_unprotected\_noncross\_turn & 275 &  \\
 & traversing\_narrow\_lane & 26 &  \\
 & behind\_long\_vehicle & 711 &  \\
 & behind\_pedestrian\_on\_driveable & 69 &  \\
 & behind\_pedestrian\_on\_pickup\_dropoff & 26 &  \\
 & behind\_bike & 205 &  \\
 & near\_multiple\_vehicles & 1363 &  \\
\multirow{-26}{*}{Hard} & traversing\_pickup\_dropoff & 3087 & \multirow{-26}{*}{16935/33.9\%} \\ \hline
 & near\_high\_speed\_vehicle & 1204 &  \\
 & waiting\_for\_pedestrian\_to\_cross & 301 &  \\
 & accelerating\_at\_crosswalk & 137 &  \\
 & accelerating\_at\_stop\_sign\_no\_crosswalk & 84 &  \\
 & accelerating\_at\_stop\_sign & 152 &  \\
 & accelerating\_at\_traffic\_light & 100 &  \\
 & accelerating\_at\_traffic\_light\_with\_lead & 115 &  \\
 & accelerating\_at\_traffic\_light\_without\_lead & 357 &  \\
 & following\_lane\_with\_lead & 214 &  \\
 & following\_lane\_with\_slow\_lead & 828 &  \\
 & on\_all\_way\_stop\_intersection & 918 &  \\
 & on\_intersection & 1142 &  \\
 & on\_stopline\_crosswalk & 1014 &  \\
 & on\_stopline\_stop\_sign & 788 &  \\
 & on\_stopline\_traffic\_light & 1342 &  \\
 & on\_traffic\_light\_intersection & 2029 &  \\
 & starting\_low\_speed\_turn & 141 &  \\
 & starting\_straight\_stop\_sign\_intersection\_traversal & 323 &  \\
 & starting\_straight\_traffic\_light\_intersection\_traversal & 364 &  \\
 & stopping\_at\_crosswalk & 219 &  \\
 & stopping\_at\_stop\_sign\_no\_crosswalk & 100 &  \\
 & stopping\_at\_stop\_sign\_without\_lead & 166 &  \\
 & stopping\_at\_traffic\_light\_with\_lead & 75 &  \\
 & stopping\_at\_traffic\_light\_without\_lead & 122 &  \\
 & stopping\_with\_lead & 219 &  \\
 & traversing\_crosswalk & 1665 &  \\
 & traversing\_intersection & 1926 &  \\
\multirow{-29}{*}{General} & traversing\_traffic\_light\_intersection & 3581 & \multirow{-29}{*}{19626/39.2\%} \\ \hline
 & stationary\_at\_crosswalk & 63 &  \\
 & stationary\_at\_traffic\_light\_with\_lead & 1206 &  \\
 & stationary\_at\_traffic\_light\_without\_lead & 1680 &  \\
 & low\_magnitude\_speed & 1788 &  \\
 & medium\_magnitude\_speed & 3547 &  \\
  & following\_lane\_without\_lead & 1388 &  \\
\multirow{-6}{*}{Simple} & high\_magnitude\_speed & 3768 & \multirow{-6}{*}{13440/26.9\%} \\ \hline
\end{tabular}
}
\end{table*}

\subsection{GRPO for VLM Driving Agents}\label{app:grpo_for_agents}
This section provides the detailed formulation of the GRPO implementation. We employ three rewards, a location reward, a velocity reward, and a format reward, for our GRPO training. The first two are calculated from the L2 error between each predicted point and its ground truth point, while the latter is consistent with that used in~\cite{deepseekr1_grpo}. Note the $K=3$ rewards as $r_1$, $r_2$, $r_3$ with their weights $\alpha_1$, $\alpha_2$, and $\alpha_3$, the normalized advantage $\mathcal{A}_i$ for an output in one group is calculated as:
\begin{equation}
\label{eq:grpo_advantage}
\textstyle
\mathcal{A}_i = \frac{R_i - \mathrm{mean}(R_1, \dots, R_G)}{\mathrm{std}(R_1, \dots, R_G)}, \quad \text{where} \quad R_i\!=\!\sum_{k=1}^{K}\!\alpha_k\!\cdot\! r_k.
\end{equation}
Besides, in equation~\ref{eq:grpo_main}, the $\tilde{w}_i$ is calculated as:
\begin{equation}
\textstyle
    \tilde{w}_i = \min\left(\frac{\pi_{\theta}(o_i|(I_j,T_j))}{\pi_{\theta_{\text{old}}}(o_i|(I_j,T_j))} , \operatorname{clip}\left(\frac{\pi_{\theta}(o_i|(I_j,T_j))}{\pi_{\theta_{\text{old}}}(o_i|(I_j,T_j))}, 1 - \epsilon, 1 + \epsilon\right) \right),
\end{equation}
where the $\epsilon$ is a tunable hyperparameter. The KL-divergence penalty item $\mathbb{D}_{\mathrm{KL}}$ is calculated as:
\begin{equation}
\textstyle
    \mathbb{D}_{\mathrm{KL}}(\pi_{\theta} || \pi_{\text{ref}}) \!=\! \frac{\pi_{\text{ref}}(o_i|(I_j,T_j))}{\pi_{\theta}(o_i|(I_j,T_j))}\! -\! \log \frac{\pi_{\text{ref}}(o_i|(I_j,T_j))}{\pi_{\theta}(o_i|(I_j,T_j))}\! -\! 1.
\end{equation}

\subsection{Tuning Ablation Details}\label{app:ablation_details}
Our tuning ablation experiments are based on our proposed dataset, DriveMind, which contains samples with modular ground truth input and detailed plan-aligned CoT reasoning towards the planning task. We conduct extensive data-driven ablations on different models, verifying the disconnect between reasoning and planning and the shortcut learning leveraging text priors in VLM driving agents. The detailed supplementary information of our ablations is shown in Table~\ref{tab:ablation_details}. \XRiii{All experiments are conducted on 4$\times$ NVIDIA A100-80G GPUs. Each SFT run takes approximately 3 days, and each GRPO run takes approximately 5 days. We note that reported times are approximate as experiments are run on a shared cluster.}

\begin{table*}[t]
\centering
\scriptsize
\setlength{\tabcolsep}{6pt}
\caption{Details of our tuning ablation analysis settings.}
\label{tab:ablation_details}
\begin{tabular}{p{0.16\textwidth} p{0.74\textwidth}}
\toprule
\textbf{Agent} & \textbf{Ablation Details with DriveMind} \\
\midrule
Base & Base model without any finetuning. Serves as a control group. \\[4pt]
CoT & Finetuned on the full DriveMind dataset with $\sim$50K CoT VQA samples. 
\\[4pt]
Plan & The reasoning part is removed from each sample in DriveMind, retaining only the final planning results. \\[4pt]
Plan\_NoV & Same as `Plan', but all visual inputs are also removed. \\[4pt]
CoT\_NoHis & Ego's history information is removed from the textual input of the samples in DriveMind. \\[4pt]
CoT\_NoHis\_Ego & Both history and the ego state are removed from the textual input of the samples in DriveMind. \\[4pt]
CoT\_NoPri & All driving priors, history, ego state, and navigation, are removed from the textual input. \\[4pt]
CoT\_L & Scalability test: Replicates the `CoT' setup on the LLaVA-1.6. \\[4pt]
Plan\_L\_NoV & Scalability test: Replicates the `Plan\_NoV' setup on the LLaVA-1.6. \\[4pt]
Omnidrive & CoT samples from the DriveMind are decomposed into sub-tasks(detection, planning, etc.) without plan-aligned reasoning. Use all the decomposed sub-task data to simultaneously train the Omnidrive agent. \\
Omnidrive\_NoPri & Same as `Omnidrive', with all the priors removed from text input for each sample. \\[4pt]
\hline \\
CoT\_grpo & The `CoT' model is further tuned with GRPO using the $1,000$ grpo VQA samples that have the same input settings as DriveMind. \\
Base\_grpo & The base model is directly tuned with GRPO using the $1,000$ grpo VQA samples. \\[4pt]
Base\_grpo\_NoPri & Same as `Base\_grpo', but all priors in the samples are removed from the text input during GRPO tuning. \\
\bottomrule
\end{tabular}
\end{table*}

\subsection{Training Details for Agents}\label{app:training-details}

In this work, we conduct standard SFT and GRPO training on our agents based on the LLM training and development framework, Scalable Lightweight Infrastructure for Fine-Tuning (SWIFT)~\cite{swift-train}. For the SFT, we employ LoRA with rank and alpha both set to 64. Training is conducted on the aligner, ViT, and LLM backbone, using learning rates of 1e-4, 1e-5, and 1e-5, respectively. The batch size is set to 8, with a warm-up ratio of 0.05. Cosine learning rate scheduling is applied to achieve smoother training. For the GRPO training, we employ the same settings, plus the temperature parameter for the policy model that generates samples is set to 0.9 to encourage a degree of exploration, the group size is set to 8, and the warm-up ratio is set to 0.01. For the three rewards, we assign weights of 0.45, 0.45, and 0.1 to the location, velocity, and format rewards for \textit{CoT\_grpo}, respectively, whereas for \textit{Base\_grpo}, the three rewards are equally weighted.

\subsection{nuPlan Challenge Metrics}\label{app:nuplan_challenge}
This section provides detailed definitions for the nuPlan Challenge~\cite{motional_nuplan_devkit_challenge} evaluation metrics used in our experiments.

\subsubsection{Open-loop metrics}
In each scenario, the agent's planned trajectory is compared against the expert's ground-truth trajectory over 1s, 2s, and 3s horizons. The comparison is based on the following five core metrics, which are first computed at each sampled timestep and then averaged over all timesteps in the scenario.

\textbf{Average Displacement Error (ADE).} At each timestep, ADE is the mean of the pointwise L2 distances between the planned and expert trajectories over the selected future horizon.

\textbf{Final Displacement Error (FDE).} At each timestep, FDE is the L2 distance between the planned and expert trajectories at the end of the selected future horizon.

\textbf{Average Heading Error (AHE).} At each timestep, AHE is the mean of the absolute differences in heading angle between the planned and expert trajectories over the future horizon.

\textbf{Final Heading Error (FHE).} At each timestep, FHE is the absolute difference in heading angle at the end of the future horizon.

\textbf{Miss Rate.} At each timestep, a ``miss'' is declared if the maximum pointwise L2 distance between the planned and expert trajectories exceeds a  threshold, which is defined as {2.0m} for the 1s horizon, {3.2m} for 2s, and {6.0m} for 3s. The scenario's Miss Rate is the ratio of ``missed'' timesteps to the total number of timesteps.

\textbf{Open-Loop Score.}
The final Open-Loop Score reported in the paper is a composite metric derived from the core metrics via a two-stage process. First, the scenario-level average of each of the five core metrics is compared against a predefined threshold to generate five binary scores ($S_{\text{metric}} \in \{0, 1\}$, where 1 represents a pass). A failure score of 0 is assigned if the scenario's Miss Rate is {greater than 30\%}, if the average ADE or FDE is {greater than 8.0 meters}, or if the average AHE or FHE is {greater than 0.8 radians}.
These binary scores are then aggregated into a final score ranging from 0 to 100. The $S_{\text{MissRate}}$ score acts as a gate; if it is 0, the final score for the scenario is 0. Otherwise, the final score is a weighted average of the other four binary scores, calculated as:
\begin{equation}
\label{eq:appendix_open_loop_score}
\textstyle
\text{Score} = \frac{2 \cdot S_{\text{ADE}} + 2 \cdot S_{\text{FDE}} + S_{\text{AHE}} + S_{\text{FHE}}}{6} \times S_{\text{MissRate}} \times 100
\end{equation}

\subsubsection{Close-loop metrics}
The Closed-Loop Score evaluates the agent's ability to drive safely, adhere to traffic rules, and maintain comfort in an interactive simulation. The score is a composite metric calculated from eight core metrics using a hybrid gating and weighted-average formula. These metrics are divided into two main groups, Gating Metrics and Weighted Metrics.

The first group consists of three {gating metrics}, which represent critical safety and progress requirements. A failure in any of these results in an immediate scenario score of 0. 

\textbf{No At-Fault Collisions (Collision Ratio).} This metric is a three-tiered score. It is assigned a value of 0 for an at-fault collision with a vehicle or Vulnerable Road User (VRU), or for multiple at-fault collisions with objects. It is 0.5 for a single at-fault collision with an object, and 1 otherwise. A collision is considered ``at-fault'' if it should have been preventable by the planner, such as colliding with a stopped object or a frontal collision.

\textbf{Drivable Area Compliance.} This is a binary score that becomes 0 if, at any frame, the maximum distance of any corner of the ego's bounding box from the nearest drivable area is more than  0.3m. This tolerance accounts for potential over-approximations of the ego bounding box.

\textbf{Making Progress.} This score is a binary value that is 0 if the agent's progress ratio along the expert route falls below the min progress threshold of 0.2.

The second group comprises five {weighted metrics}, which are scored between 0 and 1 and contribute to the final performance score via a weighted average. 

\textbf{Driving Direction Compliance.} This score is based on the distance traveled against traffic flow over a 1s time horizon; it is 1 if this distance is less than 2m, 0 if it is more than 6m, and 0.5 otherwise.

\textbf{Time to Collision (TTC).} This score is a binary value based on the minimum TTC, which is calculated by projecting bounding boxes forward up to a 3.0s horizon. The score is 0 if the minimum TTC falls below the least TTC safety threshold of 0.95s, and 1 otherwise.

\textbf{Speed Limit Compliance.} This metric evaluates the agent's adherence to posted speed limits from the map data, penalizing both the magnitude and duration of over-speeding.
This score is a continuous value between 0 and 1, calculated as: $
S_{\text{speed}} = \max\left(0, 1 - \frac{v_{\text{int}}}{v_{\text{thresh}} \cdot T_{\text{scenario}}}\right)$,
where $v_{\text{int}}$ is the time-integral of the speed violation (i.e., the area under the over-speed vs. time graph), $T_{\text{scenario}}$ is the total duration of the scenario, and $v_{\text{thresh}}$ is the maximum acceptable over-speeding threshold, set to 2.23~m/s ($\approx$~5~mph). A score of 1 indicates no violations, and the score trends towards 0 as the severity and duration of over-speeding increase.

\textbf{Ego Progress along Expert Route.} 
This metric measures how effectively the agent advances along the path taken by the expert driver.  At each timestep, the agent's movement is projected onto this path to calculate its per-frame progress. The total progress for both the agent, $P_{\text{ego}}$, and the expert, $P_{\text{expert}}$, is then calculated by integrating these per-frame values over the entire scenario.
This score is the ratio of these two values, capped between 0 and 1, and is calculated as:$
S_{\text{progress}} = \min\left(1, \frac{\max(P_{\text{ego}}, \tau)}{\max(P_{\text{expert}}, \tau)}\right)$,
where $\tau$ is a small threshold (0.1m) used to handle minor negative progress values that can arise from data noise and to prevent division by zero in scenarios with no movement. A score of 1 indicates the agent made at least as much progress as the expert.

\textbf{Comfort.}
This metric quantifies the smoothness of the planned trajectory by ensuring key variables remain within bounds representative of a comfortable human driving experience. At each timestep of the scenario, a set of quantities are checked against empirically determined thresholds derived from an analysis of expert human driver trajectories. The scenario is deemed uncomfortable, and the score $S_{\text{comfort}}$ is set to 0, if \textit{any} of the following conditions are violated at \textit{any} time: longitudinal acceleration is outside the range of [-4.05, 2.40]~m/s$^2$; absolute lateral acceleration exceeds 4.89~m/s$^2$; absolute yaw rate exceeds 0.95~rad/s; absolute yaw acceleration exceeds 1.93~rad/s$^2$; absolute longitudinal jerk exceeds 4.13~m/s$^3$; or the magnitude of the jerk vector exceeds 8.37~m/s$^3$. If none of these bounds are violated throughout the entire scenario, the score $S_{\text{comfort}}$ is 1.

\paragraph{Close-Loop Score.} The final Closed-Loop Score is the product of a {gating factor} ($S_{\text{gate}}$) and a {weighted performance score} ($S_{\text{weighted}}$). First, the  {gating factor} is the product of the three binary gating metrics, ensuring all critical safety rules are met:
\begin{equation}
\textstyle S_{\text{gate}} = S_{\text{no\_collision}} \times S_{\text{drivable\_area}} \times S_{\text{making\_progress}}
\end{equation}
Second, the weighted performance score is the weighted average of the five weighted metrics, with a total weight of 21:
\begin{equation}
\textstyle S_{\text{weighted}} = \frac{5S_{\text{direction}} + 5S_{\text{TTC}} + 4S_{\text{speed}} + 5S_{\text{progress}} + 2S_{\text{comfort}}}{21}
\end{equation}
The final Closed-Loop Score is then calculated as follows, where if $S_{\text{gate}}$ is 0, the total score is 0:
\begin{equation}
\label{eq:appendix_closed_loop_score}
\textstyle \text{Score} = S_{\text{gate}} \times S_{\text{weighted}} \times 100
\end{equation}

\subsection{Sequence-level Segment Attention Analysis for Omnidrive}
\begin{table*}[t]
\centering
\scriptsize 
\renewcommand{\arraystretch}{1.05} 
\setlength{\tabcolsep}{6pt} 
\caption{
    Attention distributions for the \textit{Omnidrive} across preceding sequences during target sequence generation.
}
\label{tab:attn_omnidrive}
\begin{tabular}{@{}lcccccc@{}}
\toprule
\multirow{2}{*}{\textbf{Attention Target}} & \multicolumn{3}{c}{\textbf{Detection Task}} & \multicolumn{3}{c}{\textbf{Planning Task}} \\
\cmidrule(lr){2-4} \cmidrule(lr){5-7}
& Shallow Layer& Middle Layer& Final Layer& Shallow Layer& Middle Layer& Final Layer \\
\midrule
Image Tokens (\%) & \textbf{3.46} & \textbf{10.02} & \textbf{15.92} & 1.96 & 3.22 & 4.05 \\
Textual Priors (\%) & 6.76 & 5.93 & 6.89 & \textbf{10.52} & \textbf{9.93} & \textbf{23.97} \\
Textual Tokens (\%) & 96.54 & 89.98 & 84.08 & 98.04 & 96.78 & 95.95 \\
\bottomrule
\end{tabular}
\end{table*}

We also conduct the sequence-level attention analysis on the \textit{Omnidrive} model. We primarily select the object detection task (reflecting implicit reasoning processes) and the planning task from the decomposed CoT subtasks. Since there is no explicit reasoning process, we mainly examine the attention distribution between the sequence of image tokens and the sequence of textual prior tokens in the input across different depth layers during model output.

The results in Table~\ref{tab:attn_omnidrive} demonstrate that \textit{Omnidrive} progressively focuses more on the image sequence during detection tasks (implicit reasoning). At the final layer, its attention to image tokens reaches 15.92\%, while attention to the textual prior sequence fluctuates stably between approximately 6-7\%, significantly lower than that for image tokens. This indicates that the textual prior component does not play a decisive role in the detection task that provides the implicit reasoning ability.

During planning tasks, \textit{Omnidrive} consistently allocates significantly high attention weights to textual prior sequences across different layers, ultimately reaching approximately 24\%, which is much higher than the attention weights to the image sequence. This demonstrates that \textit{Omnidrive} also heavily relies on textual prior information for planning. The attention weights assigned to the image sequence by \textit{Omnidrive} in the planning task also increase slightly as the layer depth increases. This may stem from the absence of preceding reasoning sequences compared to \textit{CoT\_grpo}, resulting in fewer tokens being processed, and consequently increasing attention to images. It may also reflect the model's implicit, superficial and limited processing of image information, as the final attention weight only accounts for 4\%. 

The results of the sequence-level attention analysis on \textit{Omnidrive} further support our conclusion: current VLM driving agents heavily rely on textual priors during planning, with reasoning processes being almost irrelevant. This also shows that counterfactual reasoning fails to resolve this issue. Furthermore, results also show that modality bias in comprehension persists within \textit{Omnidrive}.

\subsection{More Examples of Causal Probe}\label{app:examples-of-causal-probe}

This section provides more examples of our causal probe applied to diverse driving scenarios, further illustrating the reasoning-planning disconnect and the efficacy of our diagnostic method.

\begin{figure*}[t] \centering
\includegraphics[width=\textwidth,height=0.15\textwidth]{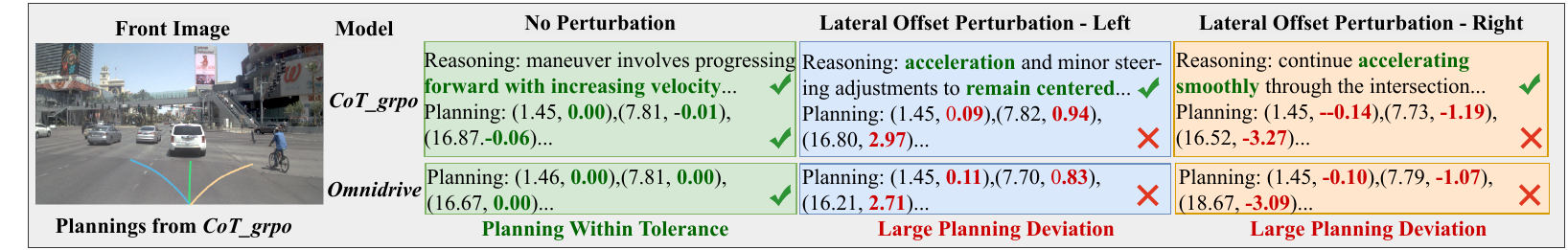}
    \caption{
   An illustration of our causal probe's diagnostic capabilities in a straight driving scenario. The figure contrasts an agent's original plan with its divergent trajectories when subjected to lateral offset perturbations to the left and right. The lateral trajectories of both models' output undergo a significant change in response to the perturbations.}
    \label{fig:calsual_prob_app_straight}
\end{figure*}

\begin{figure*}[t] \centering
\includegraphics[width=\textwidth,height=0.15\textwidth]{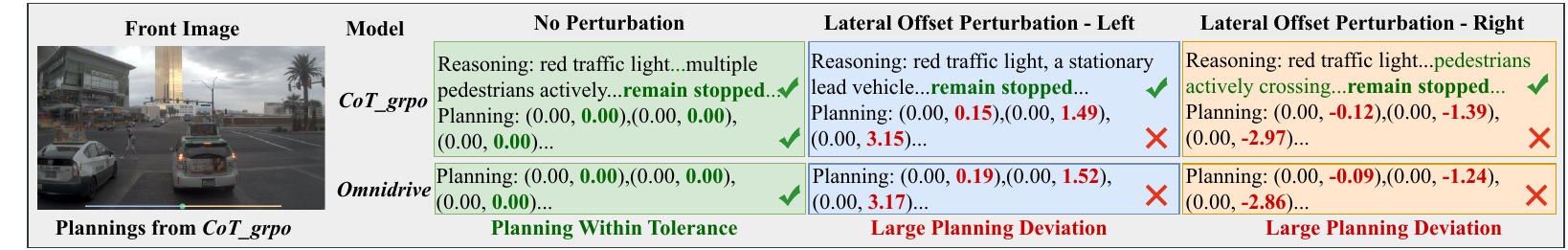}
    \caption{
   An illustration of our causal probe's diagnostic capabilities in a stopping scenario. The figure contrasts an agent's original plan with its divergent trajectories when subjected to lateral offset perturbations to the left and right. Models output distinct lateral ‘translation’ trajectories after perturbation, fully demonstrating the disconnect between reasoning and planning, as well as the existence of shortcut learning dependent on text priors.}
    \label{fig:calsual_prob_app_stop}
\end{figure*}

Figure~\ref{fig:calsual_prob_app_straight} illustrates the application of the causal probe in a simple straight driving scenario. In the baseline case with no perturbation, both the \textit{CoT\_grpo} and \textit{Omnidrive} agents correctly produce a plan to continue straight. However, when a minor lateral offset is introduced into the textual priors, both agents exhibit catastrophic planning failures. Their planned trajectories deviate significantly to the left or right, directly following the perturbed history.
Crucially, for the \textit{CoT\_grpo} agent, the generated reasoning remains correct and consistent across all three cases (e.g., ``Reasoning: continue accelerating and minor steer''). Despite the correct reasoning, the planning module produces a wildly divergent and unsafe trajectory. This provides a clear, qualitative example of the planning module ignoring the reasoning module and instead relying on the shortcut provided by the textual priors.

Figure~\ref{fig:calsual_prob_app_stop} provides an even more stark example of disconnect in a stationary scenario. In the baseline case, both agents correctly plan to remain stopped at a red light while pedestrians are present. Their reasoning correctly identifies the red light and the need to wait. However, when a lateral offset is applied to the velocity  priors, both agents generate an unsafe plan that involves a significant and unnecessary lateral swerve, even while stationary.
This case is particularly revealing. The reasoning module for \textit{CoT\_grpo} continues to correctly state that the agent should remain stopped due to the red light and pedestrians. Yet, the planning module, completely decoupled from this reasoning, still reacts to the perturbed prior and outputs an erroneous, non-zero motion plan. This demonstrates how deeply ingrained the shortcut learning is, as it can override even a fundamental safety behavior like remaining stopped at a red light.

\subsection{\rebuttal{Evaluation by Scenario Complexity}}
\rebuttal{To validate the consistency of our findings across different scene complexities, we stratify the 14 distinct evaluation types in the nuPlan challenge test set to the hard, general and simple categories, as stated in Appendix~\ref{app:dataset_proportion_details}. We measure the performance of the standard agent (\textit{CoT}), the blind agent (\textit{Plan\_NoV}), and the prior-deprived agent (\textit{CoT\_NoPri}) across these categories. The evaluation results of the 3s planning are summarized in Table~\ref{tab:stratified_scenario_scores}.

As observed in the table, the ``blind" \textit{Plan\_NoV} agent achieves parity with (or slightly exceeds) the multimodal \textit{CoT} agent across all complexity levels, including ``Hard" scenarios. This confirms that the reasoning-planning disconnect is not limited to simple cases; even in complex turning scenarios, the model ignores visual cues in favor of priors. The performance gap between the standard agent and the prior-removed agent ($\Delta$ (\textit{CoT} - \textit{CoT\_NoPri})) serves as a proxy for ``reliance on priors." Notably, this gap widens significantly as complexity increases: the Open-Loop score of \textit{CoT\_NoPri} drops by 47.16 points in Hard scenarios compared to only 18.50 points in Simple scenarios. These results confirm that the model's reliance on shortcuts is not only consistent but also effectively strengthened under challenging conditions. Rather than engaging in deeper visual reasoning for complex tasks, the agent leans more heavily on textual priors to solve the problem.
}

\begin{table*}[t]
    \centering
    \footnotesize
    \renewcommand{\arraystretch}{1.3} 
    \setlength{\tabcolsep}{2pt}       

    \caption{Open-loop and closed-loop performance comparison across scenario categories.}
    \label{tab:stratified_scenario_scores}

    \begin{tabularx}{\textwidth}{l >{\raggedright\arraybackslash}X ccc ccc}
        \toprule
        \multirow{2.5}{*}{\textbf{Category}} & 
        \multirow{2.5}{*}{\textbf{Scenario}} & 
        \multicolumn{3}{c}{\textbf{Open-Loop Score $\uparrow$}} & 
        \multicolumn{3}{c}{\textbf{Closed-Loop Score $\uparrow$}} \\
        \cmidrule(lr){3-5} \cmidrule(lr){6-8} 
        
        & & 
        \textbf{CoT} & \textbf{CoT\_NoPri} & \textbf{Plan\_NoV} & 
        \textbf{CoT} & \textbf{CoT\_NoPri} & \textbf{Plan\_NoV} \\
        \midrule

        \multirow{7}{*}{\textbf{Hard}} 
        & high\_lateral\_acceleration & 89.69 & 40.74 & 90.04 & 90.31 & 66.18 & 90.51 \\
        & behind\_long\_vehicle       & 98.31 & 67.25 & 98.57 & 99.28 & 77.74 & 99.45 \\
        & changing\_lane              & 92.33 & 27.48 & 92.40 & 85.96 & 67.45 & 86.08 \\
        & starting\_left\_turn        & 86.98 & 36.94 & 87.26 & 82.93 & 69.91 & 83.41 \\
        & starting\_right\_turn       & 88.64 & 42.56 & 88.94 & 84.63 & 68.14 & 85.49 \\
        & near\_multiple\_vehicles    & 93.42 & 42.72 & 93.51 & 92.79 & 81.82 & 92.67 \\
        & traversing\_pickup\_dropoff & 89.80 & 51.35 & 89.99 & 84.03 & 69.27 & 84.26 \\
        \addlinespace[2pt] 
        \multicolumn{2}{l}{\textbf{Hard Avg}} & \textbf{91.31} & \textbf{44.15} & \textbf{91.53} & \textbf{88.56} & \textbf{71.50} & \textbf{88.84} \\
        \rowcolor{gray!20}
        \multicolumn{2}{l}{\textit{$\Delta$ (CoT - CoT\_NoPri)}} & &  \textbf{+47.16} & &  & \textbf{+17.06} &  \\
        \midrule

        \multirow{4}{*}{\textbf{General}} 
        & waiting\_for\_pedestrian\_ \newline to\_cross & 90.14 & 44.68 & 90.81 & 95.27 & 82.76 & 95.06 \\
        \addlinespace[2pt] 
        & starting\_straight\_traffic\_ \newline light\_intersection\_traversal & 91.84 & 42.78 & 92.09 & 93.57 & 81.49 & 93.26 \\
        \addlinespace[2pt]
        & stopping\_with\_lead        & 93.03 & 67.84 & 94.87 & 94.14 & 81.75 & 94.35 \\
        & following\_lane\_with\_lead & 95.16 & 67.07 & 96.14 & 99.98 & 91.49 & 99.98 \\
        \addlinespace[2pt]
        \multicolumn{2}{l}{\textbf{Gen. Avg}} & \textbf{92.54} & \textbf{55.59} & \textbf{93.48} & \textbf{95.74} & \textbf{84.37} & \textbf{95.66} \\
       \rowcolor{gray!20}
        \multicolumn{2}{l}{\textit{$\Delta$ (CoT - CoT\_NoPri)}} &  & \textbf{+36.95} & &  &\textbf{+11.37} & \\
        \midrule

        \multirow{3}{*}{\textbf{Simple}} 
        & low\_magnitude\_speed   & 92.71 & 68.73 & 92.61 & 95.26 & 85.24 & 95.37 \\
        & stationary\_in\_traffic & 99.29 & 94.47 & 99.08 & 99.06 & 97.74 & 98.93 \\
        & high\_magnitude\_speed  & 94.55 & 67.86 & 94.56 & 96.13 & 87.72 & 96.78 \\
        \addlinespace[2pt]
        \multicolumn{2}{l}{\textbf{Simple Avg}} & \textbf{95.52} & \textbf{77.02} & \textbf{95.42} & \textbf{96.82} & \textbf{90.23} & \textbf{97.03} \\
        \rowcolor{gray!20}
        \multicolumn{2}{l}{\textit{$\Delta$ (CoT - CoT\_NoPri)}} &&  \textbf{+18.50}  &&  & \textbf{+6.58}&  \\
        \bottomrule
    \end{tabularx}
\end{table*}

\XRiii{
\subsection{Testing on Extended Training Setups}\label{app:extended_other_exps}
In this section, we test further training setups to demonstrate that the identified reasoning-planning disconnect is fundamental to existing SFT and RL training, rather than an artifact.
\subsubsection{Supervised Fine-tuning: Sequential Training}
In our primary evaluation, the \textit{CoT} agent is trained using a joint Supervised Fine-Tuning (SFT) objective, where the model learns to generate both the reasoning steps and the final plan simultaneously given the input (i.e., $\text{Input} \rightarrow [\text{CoT}, \text{Plan}]$). To investigate whether the observed reasoning-planning disconnect is merely an artifact of this joint training paradigm, we conduct a two-stage sequential training experiment.

In this experiment, we sequentially train two agents with and without text priors, denoted as \textit{CoT\_Seq} and \textit{CoT\_Seq\_NoPri}, using the following two-stage process:
\begin{itemize}
    \item[] \textbf{Stage 1 (Reasoning Generation):} The model is trained solely to generate the reasoning steps given the input ($\text{Input} \rightarrow \text{CoT}$).
    \item[] \textbf{Stage 2 (Planning Generation):} A planning module is trained using the input and the model's \emph{own generated} reasoning from Stage 1 to predict the final trajectory ($[\text{Input}, \text{Model's\_Own\_CoT}] \rightarrow \text{Plan}$).
\end{itemize}
The results are presented in Table~\ref{tab:sequential_results}. The performance patterns of the sequential agents are similar to the joint SFT setting. The \textit{CoT\_Seq} agent achieves high scores, but its planning ability drops drastically when priors are removed (\textit{CoT\_Seq\_NoPri}), confirming that this two-stage training approach still fails to break the reliance on textual priors. The disconnect is thus fundamental to the existing SFT training or VLM architectural vulnerability, regardless of whether the tasks are learned jointly or sequentially.
}
\begin{table*}[t]
\centering
\caption{Planning performance (3s) of agents trained with the sequential SFT setup. Removing priors, \textit{CoT\_Seq\_NoPri} still exhibits a substantial performance degradation.}
\label{tab:sequential_results}
\begin{tabular}{lcc}
\toprule
\textbf{Agent} & \textbf{Open-Loop Score $\uparrow$} & \textbf{Close-Loop Score $\uparrow$} \\
\midrule
\textit{CoT\_Seq}       & 91.64 & 91.45 \\
\textit{CoT\_Seq\_NoPri}& 49.61 & 74.62 \\
\bottomrule
\end{tabular}
\end{table*}

\XRiii{
\subsubsection{RL: GRPO Training with CoT Rewards}
In our previous Reinforcement Learning (RL) setup using Group Relative Policy Optimization (GRPO), we primarily employ a `format' reward. This reward imposes structural discipline on the reasoning process, forcing the model to engage the Chain-of-Thought (CoT) module and adhere to a required structural sequence (e.g., \texttt{<think>} $\rightarrow$ \texttt{<answer>}). Though this implicitly encourages the model to analyze the scene from input to output and standardizes the trajectory extraction, here we design and add a more detailed reward $R_{\text{visual\_match}}$, which scores the alignment of the generated CoT with key visual components in the scene:
\begin{equation}
    R_{\text{visual\_match}} = \frac{\text{count}(\text{key\_visual\_components\_in\_CoT})}{\text{count}(\text{all\_key\_visual\_components})},
\end{equation}
where `key visual components' are concretely defined as the critical environmental elements extracted from the ground truth that directly influence driving decisions. Specifically, these include active traffic lights, relevant road signs, and dynamic obstacles (such as pedestrians or vehicles) within a 20m radius. This is to test whether a different reward, such as $R_{\text{visual\_match}}$ that directly scores the alignment between the CoT and visual evidence, and a better CoT would mitigate the disconnect.

We train two new agents, \textit{Base\_grpo\_visual\_match} and \textit{CoT\_grpo\_visual\_match}, with this explicitly grounded reward. Both agents successfully learn to articulate visual facts, achieving high average matching scores greater than 0.7. However, we then evaluate these agents under our causal probe ($\delta = 0.1$) to test genuine scene-grounded planning. The results are summarized in Table~\ref{tab:grpo_visual_match}.

\begin{table*}[t]
\centering
\caption{Performance of agents trained with the visual matching reward under the causal probe ($\delta = 0.1$). Despite high visual matching scores during training, both agents exhibit large deviations.}
\label{tab:grpo_visual_match}
\begin{tabular}{lcc}
\toprule
\textbf{Agent} & \textbf{MFD $\downarrow$} & \textbf{RID $\downarrow$} \\
\midrule
\textit{CoT\_grpo\_visual\_match}  & 5.77 & 1.99 \\
\textit{Base\_grpo\_visual\_match} & 4.42 & 1.61 \\
\bottomrule
\end{tabular}
\end{table*}

As shown in the table, both driving agents still exhibit massive deviations. Even when the model is successfully incentivized to perceive and accurately articulate correct visual facts reasoning (e.g., correctly identifying a red light in text), the planning module still bypasses this reasoning to rely on textual priors. These results show that the reasoning-planning disconnect is a fundamental issue during current RL training, merely a lack of visual attention supervision nor low-quality CoT.

In summary, the reasoning-planning disconnect is a fundamental issue with current VLM driving agents' training. Exploring new SFT paradigms and developing advanced causality rewards to genuinely rebuild the reasoning-planning link remains a critical path for future research. We view those as orthogonal to our current work, and we believe our reasoning-planning decoupling hypothesis and probing tools we have developed are the essential groundwork that future solution-oriented research must build upon.

}

\subsection{Broader Impact Statement}\label{app:impact statement}
\XRiii{This work identifies a critical safety concern in VLM-based driving agents: their planning may not be causally grounded in reasoning despite appearing interpretable. By exposing this disconnect and providing diagnostic tools, we aim to promote more robust development and evaluation practices for autonomous driving systems before real-world deployment. We think there are no negative societal impacts of this diagnostic research.}

\subsection{The Use of Large Language Models}
\label{app:use_of_LLM_declaration}
In preparing this manuscript, we utilized an LLM (i.e., GPT-5) as a tool to enhance the clarity and precision of our writing. The core ideas, arguments, and scientific contributions are conceived and authored by the human writers. The LLM was employed to refine sentence structure, improve readability, and ensure grammatical correctness on initial human-written drafts. The final version of the manuscript reflects the authors' thorough review and final judgment on all text.

\end{document}